\documentclass[11pt]{article}

\usepackage[final]{acl}

\usepackage{times}
\usepackage{latexsym}
\usepackage[T1]{fontenc}
\usepackage[utf8]{inputenc}
\usepackage{microtype}
\usepackage{inconsolata}
\usepackage{url}

\usepackage{graphicx}
\usepackage{multicol}
\usepackage{placeins}
\usepackage{booktabs}
\usepackage{multirow}
\usepackage{amsmath}
\usepackage{enumitem}
\usepackage{pifont}
\AtBeginDocument{\setlength{\textfloatsep}{6pt plus 1pt minus 1pt}}
\usepackage{tcolorbox}
\tcbuselibrary{breakable}
\newenvironment{promptbox}[1]{%
    \begin{tcolorbox}[breakable, title={#1}]
}{%
    \end{tcolorbox}
}

\definecolor{darkblue}{rgb}{0, 0, 0.5}
\hypersetup{colorlinks=true, citecolor=darkblue, linkcolor=darkblue, urlcolor=darkblue}

\title{How Well Do LLMs Reason with Noisy Evidence? An Active Visual Reasoning Benchmark}

\author{
  \textbf{Bach Nguyen} \ \
  \textbf{Zhaonan Li} \ \
  \textbf{Mau Son Nguyen} \ \
  \textbf{Sanika Chavan} \\
  \textbf{Nilay Kumar} \ \
  \textbf{Hong Anh Nguyen} \ \
  \textbf{Khoa Vo} \ \
  \textbf{Ben Zhou} 
\\[4pt]
  Arizona State University
}

\newcommand{\method}[0]{\textsc{VisualNoiseQA}}

\begin{document}

\maketitle

\begin{abstract}

Real-world reasoning rarely reduces to static question answering: agents must actively gather information from tools and sensors that are often noisy and unreliable. Yet most existing active reasoning benchmarks assume that environmental feedback is trustworthy, or introduce noise without exposing an explicit, calibrated uncertainty signal, leaving open how LLMs should reason when the evidence itself is uncertain. We introduce \method{}\footnote{We will release all code and data under an open-source license upon publication.}, a novel benchmark for active reasoning under noisy visual feedback. A text-only LLM must solve VQA problems by iteratively querying a fixed, off-the-shelf VLM treated as a stochastic visual sensor. For each query, we draw multiple samples and expose an empirical uncertainty signal via self-consistency, enabling the reasoner to probe from different angles and decide what to ask next and when to stop. Our construction is automatic and scalable: starting from diverse VQA sources and two noisy VLMs, we retain only questions where the sensor is inconsistent yet human-solvable. We evaluate multiple LLM reasoners on 1,000 instances spanning perception, chart understanding, and knowledge-intensive reasoning. \method{} thus provides a controlled playground to study how different LLMs exploit uncertainty signals for robust reasoning.

\end{abstract}

\section{Introduction}

\begin{figure*}[t]
  \centering
  \includegraphics[width=\textwidth, clip]{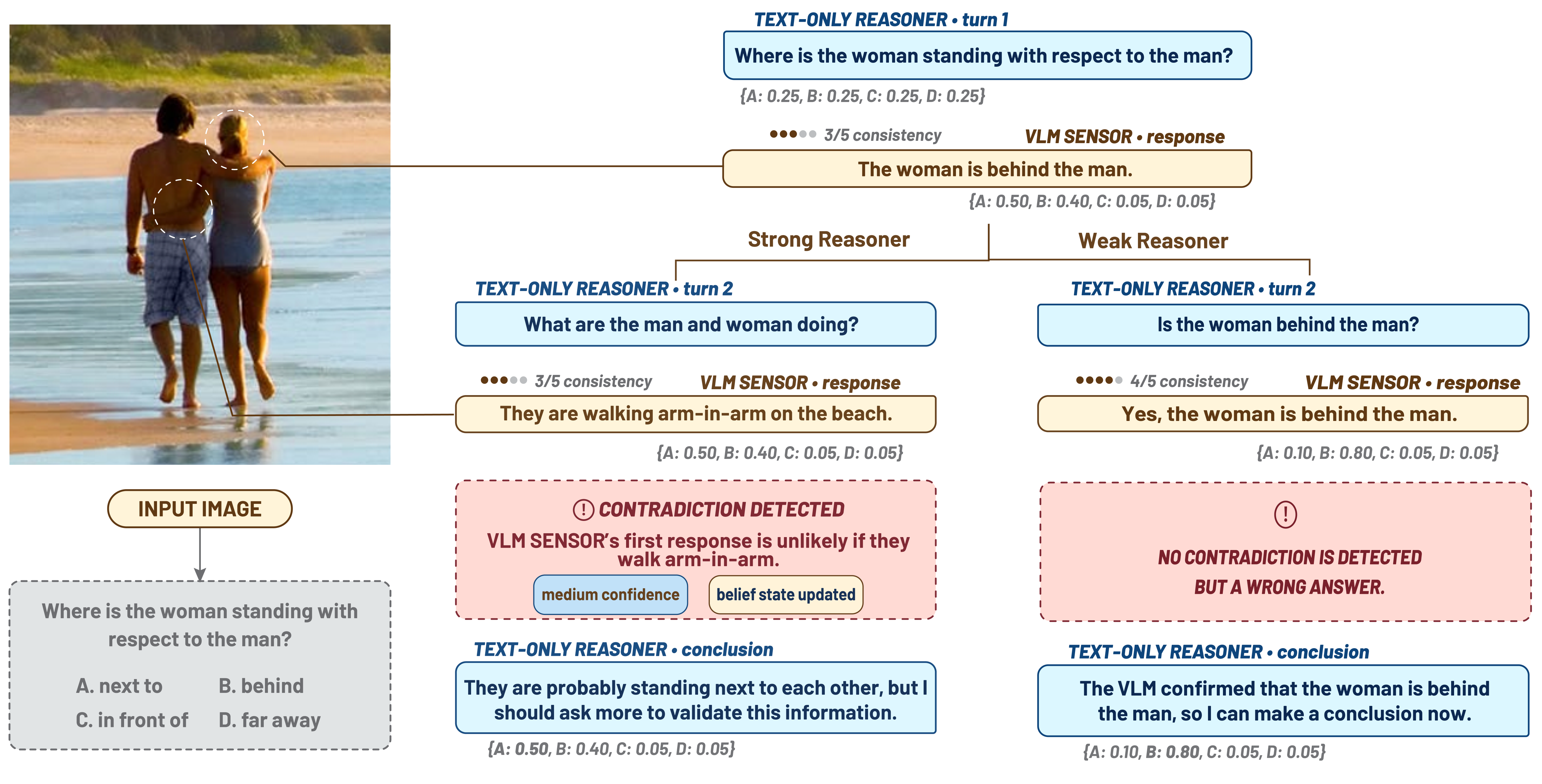}
  \vspace{-0.5cm}
  \caption{\textbf{Active reasoning under noisy visual feedback}. From the 3/5 claim ``the woman is behind the man,'' a strong reasoner diversifies its query, detects the ``arm-in-arm'' contradiction,
 and defers commitment, while a weak reasoner asks a confirmatory question and commits to a highly consistent (4/5) but wrong answer. \method{} tests whether LLMs can probe, detect contradictions, and reason under unreliable evidence.}
  \label{fig:teaser}
\end{figure*}

Humans typically solve problems by \emph{actively} gathering information from tools and sensors that are noisy and sometimes wrong: we form expectations about the world, compare new observations against these beliefs, and cross-check conflicting evidence before committing to a decision. Recent work on LLM agents and active reasoning begins to move in this direction, letting models plan tool calls, ask follow-up questions, or navigate simulators instead of answering in a single shot \citep{shridhar2020alfworld, nakano2021webgpt, yao2022webshop,abdulhai2023lmrl, li2024mediq, hu2024uncertainty, zhou2025passive}. However, these settings usually assume that the environment behaves as an oracle, or they provide no explicit and reliable signal about the uncertainty of tool outputs, leaving open how LLMs should act when feedback itself is unreliable. People, in contrast, rarely trust a single observation at face value, especially when they are explicitly asked to double-check or when the stakes are high. For instance, when a navigation app suggests an exit that conflicts with road signs, we slow down and cross-check. A similar pattern arises in vision: consider the image in Figure~\ref{fig:teaser} and the question \emph{``Where is the woman standing with respect to the man?''} A visual-language model repeatedly answers that the woman is \emph{behind} the man, yet also describes them as walking \emph{arm-in-arm}. Without seeing the image, a human reasoner can detect the inconsistency, realize that walking arm-in-arm while one person is behind the other is unlikely, and infer that they are most plausibly standing next to each other, while a weaker reasoner may instead re-ask a confirmatory question and commit prematurely to the wrong answer (Figure~\ref{fig:teaser}). This illustrates that arriving at the correct conclusion depends not only on the confidence with which each observation is reported, but also on critically assessing the reliability of each piece of evidence---particularly when responses are contradictory. This form of reliability-aware, belief-driven cross-validation is largely absent from current LLM benchmarks. 

In this paper, we study how \emph{text-only} LLMs perform active reasoning when their only access to an image is uncertain feedback from a fixed VLM sensor. Rather than pushing the frontier of visual recognition, we treat an off-the-shelf VLM as a black-box noisy sensor and use its self-consistency across samples as a cheap, observable, scalable, and externally verifiable uncertainty signal that the reasoner can act on directly. Building on prior ``blind'' VQA setups~\citep{li2025unbiased}, the LLM never sees the image: it must solve each instance by deciding what perception questions to ask, how to aggregate the answers, and when to trust or override them. For each sensor query we draw multiple VLM samples and expose a coarse consistency score, enabling the reasoner to probe from different angles and recover from misleading feedback. To study this systematically, we introduce \method{}, a benchmark and analysis suite for active reasoning under noisy visual feedback. Crucially, \method{} is designed to evaluate \emph{text-only} LLM reasoners, not vision models, isolating the ability to reason under uncertainty from visual perception itself. \method{} is constructed automatically from diverse VQA sources and two modern VLMs with distinct failure modes, selecting instances where the sensors are self-inconsistent, yet ``blind'' human annotators can still solve the task. 

We use \method{} to evaluate a range of LLM reasoners and interaction strategies, finding that all models still lag behind humans, but strong reasoners improve over direct VLM predictions and leverage consistency signals to reduce miscalibration, especially on perception tasks. These results highlight both the promise and current limitations of uncertainty-aware reasoning with noisy sensors. Beyond accuracy, our analysis surfaces failure modes that separate strong from weak reasoners and provide targets for models that reason reliably under noisy evidence. While other benchmarks and datasets focus on gathering missing information and treating received information as essentially true, our main contribution is to evaluate how well LLMs can navigate through unreliable and untrustworthy signals, which represent real-world scenarios. 

\section{Related Work}
Our work connects to modular visual reasoning frameworks that separate perception from reasoning, to LLM-based agents for active information seeking and tool use, and to RAG systems that use external sources for question answering.

\noindent \textbf{Modular visual reasoning.} 
Classical modular VQA approaches separate perception from reasoning by composing neural modules or executing programs over scene graphs and other structured visual abstractions, as in neural-symbolic VQA systems~\citep{yi2018neural}. More recent work such as ViperGPT and VisProg uses LLMs to synthesize code or programs that call visual tools while keeping a clean interface between text-based reasoning and perception~\citep{suris2023vipergpt, gupta2023visual}. Closest to our setup, recent ``blind'' VQA frameworks (e.g., \citealp{li2025unbiased}) let a text-only reasoner query a VLM sensor through restricted perception primitives; our benchmark follows this separation but focuses explicitly on \emph{noisy} VLM feedback, shifting the challenge from information gathering to reasoning under unreliable evidence.

\noindent \textbf{LLM agents and tool use.}
A growing line of work studies LLM agents that interact with external tools or environments, such as web browsers, simulators, or APIs. Early systems like WebGPT train models to browse and cite web resources for question answering~\citep{nakano2021webgpt}. ReAct interleaves chain-of-thought with actions, enabling LLMs to plan and execute multi-step tool calls in textual environments~\citep{yao2023react}, while Toolformer learns to call APIs or modular experts when needed~\citep{schick2023toolformer}. More recent agent frameworks (e.g., Voyager and other long-horizon agents) extend this paradigm to embodied and open-ended environments with persistent state and skills~\citep{wang2023voyager}. In most of these settings, tools are treated as approximately reliable oracles (e.g., a browser or API returning accurate results), whereas in our benchmark the only tool for accessing the image is an explicitly noisy and stateless VLM sensor, which provides explicit consistency information associated with each query.

\noindent \textbf{Retrieval-augmented generation (RAG).}
RAG systems augment LLMs with retrieved evidence that can itself be incomplete or conflicting~\citep{lewis2020retrieval}. Nevertheless, most RAG pipelines condition on retrieved passages as if they were reliable. \method{} can be seen as a perceptual analogue that makes this unreliability \textbf{explicit}: the reasoner queries a VLM for multiple, possibly inconsistent descriptions and must aggregate them using an exposed self-consistency signal, rather than trusting any single retrieval.

\begin{table*}[!t]
\centering
\small
\setlength{\tabcolsep}{2pt}       
\resizebox{\textwidth}{!}{    

\begin{tabular}{l l
                cc cc cc
                cc cc cc}
\toprule
& &
\multicolumn{6}{c}{InternVL3.5-8B} &
\multicolumn{6}{c}{Qwen2.5-VL-7B} \\
\cmidrule(lr){3-8}\cmidrule(lr){9-14}
& &
\multicolumn{2}{c}{Perception} &
\multicolumn{2}{c}{Charts} &
\multicolumn{2}{c}{Knowledge} &
\multicolumn{2}{c}{Perception} &
\multicolumn{2}{c}{Charts} &
\multicolumn{2}{c}{Knowledge} \\
Reasoner & Strategy &
Acc $\uparrow$ & ECE $\downarrow$ &
Acc $\uparrow$ & ECE $\downarrow$ &
Acc $\uparrow$ & ECE $\downarrow$ &
Acc $\uparrow$ & ECE $\downarrow$ &
Acc $\uparrow$ & ECE $\downarrow$ &
Acc $\uparrow$ & ECE $\downarrow$ \\
\midrule
Human & - & 72.0 & - & 86.0 & - & 69.3 & - & 70.0 & - & 92.0 & - & 74.0 & -\\
\midrule
& BP & \textbf{49.0} & - & 32.0 & - & 60.7 & - & 42.0 & - & 29.0 & - & 61.3 & - \\
GPT-5 Mini & UoT & 46.0 & - & 36.0 & - & 50.0 & - & 39.0 & - & 30.0 & - & 58.0 & -\\
& ReAct & 36.0 & - & 46.0 & - & 56.7 & - & 42.0 & - & 35.0 & - & 63.0 & - \\
\midrule[0.2pt]
& BP & 48.6 & 0.27 & \textbf{59.8} & 0.14 & \textbf{61.0} & 0.12 & 50.8 & 0.22 & \textbf{60.6} & 0.18 & \textbf{67.7} & 0.09 \\
Qwen3.5-122B-A10B & UoT & 40.6 & 0.14 & 29.0 & 0.55 & 54.3 & 0.14 & 45.2 & 0.22 & 35.4 & 0.47 & 59.9 & 0.13\\
& ReAct & 46.0 & 0.32 & 46.2 & 0.31 & 60.7 & 0.14 & \textbf{53.6} & 0.20 & 45.0 & 0.39 & 65.9 & 0.12 \\
\midrule[0.2pt]
& BP & 43.0 & 0.29 & 36.2 & 0.49 & 49.3 & 0.18 & 49.6 & 0.19 & 35.2 & 0.53 & 54.1 & 0.17 \\
GPT-oss-20B & UoT & 36.4 & 0.19 & 30.2 & 0.57 & 42.7 & 0.13 & 41.8 & 0.20 & 28.4 & 0.59 & 45.8 & 0.15 \\
& ReAct & 39.0 & 0.32 & 33.4 & 0.52 & 50.6 & 0.19 & 47.6 & 0.28 & 42.8 & 0.38 & 52.0 & 0.21 \\
\midrule[0.2pt]
& BP & 43.6 & 0.31 & 43.4 & 0.41 & 42.7 & 0.22 & 49.4 & 0.16 & 45.0 & 0.41 & 48.1 & 0.20 \\
Qwen2.5-72B & UoT & 39.2 & 0.27 & 28.4 & 0.52 & 38.9 & 0.23 & 41.8 & 0.29 & 31.2 & 0.51 & 45.0 & 0.18 \\
& ReAct & 44.0 & 0.33 & 38.4 & 0.45 & 46.7 & 0.21 & 46.6 & 0.34 & 47.0 & 0.32 & 48.9 & 0.21 \\
\midrule[0.2pt]
& BP & 39.8 & 0.23 & 38.2 & 0.34 & 43.2 & 0.16 & 43.8 & 0.14 & 40.8 & 0.30 & 45.0 & 0.20 \\
Qwen3-32B & UoT & 29.2 & 0.16 & 30.8 & 0.35 & 29.9 & 0.11 & 26.0 & 0.15 & 34.8 & 0.24 & 31.3 & 0.18 \\
& ReAct & 37.6 & 0.31 & 39.2 & 0.41 & 44.7 & 0.13 & 45.4 & 0.21 & 48.0 & 0.26 & 48.7 & 0.17 \\
\midrule[0.2pt]
& BP & 38.8 & 0.22 & 28.6 & 0.40 & 32.1 & 0.18 & 38.4 & 0.23 & 30.8 & 0.37 & 38.5 & 0.18 \\
Qwen3-8B & UoT & 27.0 & 0.20 & 19.6 & 0.42 & 24.3 & 0.15 & 27.4 & 0.23 & 20.6 & 0.45 & 26.3 & 0.18 \\
& ReAct & 38.2 & 0.31 & 25.8 & 0.60 & 36.5 & 0.22 & 46.0 & 0.19 & 29.4 & 0.57 & 39.5 & 0.24 \\
\midrule[0.2pt]
& BP & 38.0 & 0.32 & 34.8 & 0.25 & 32.9 & 0.28 & 42.8 & 0.19 & 29.2 & 0.28 & 35.7 & 0.21 \\
Qwen2.5-7B & UoT & 32.4 & 0.24 & 25.8 & 0.38 & 31.1 & 0.24 & 35.0 & 0.26 & 26.8 & 0.42 & 31.9 & 0.23 \\
& ReAct & 37.0 & 0.31 & 32.8 & 0.55 & 34.9 & 0.25 & 45.0 & 0.26 & 27.8 & 0.59 & 39.5 & 0.27 \\
\midrule
& BP & 33.8 & 0.28 & 42.2 & 0.27 & 32.7 & 0.20 & 36.8 & 0.29 & 40.8 & 0.25 & 37.3 & 0.16 \\
LLaMa3.1-8B & UoT & 32.2 & 0.25 & 32.6 & 0.47 & 32.9 & 0.16 & 34.8 & 0.20 & 32.4 & 0.45 & 30.1 & 0.23 \\
& ReAct & 38.0 & 0.28 & 51.4 & 0.17 & 34.3 & 0.18 & 46.4 & 0.18 & 58.2 & 0.16 & 36.3 & 0.22 \\
\midrule
VLM & - & 27.0 & 0.44 & 35.7 & 0.50 & 25.2 & 0.43 & 20.3 & 0.49 & 33.7 & 0.43 & 25.3 & 0.33 \\
\bottomrule
\end{tabular}
}
\caption{
Accuracy and calibration on three task categories:
\textbf{Perception} (SeedBench), \textbf{Charts} (FigureQA),
and \textbf{Knowledge} (macro-average over MMStar, CVQA, and ReasonVQA).
Full table with standard deviations across trials is reported in App.~\ref{sec:full-results} 
}

\label{tab:category-results}
\end{table*}

\noindent \textbf{Active reasoning and information seeking under uncertainty.}
Active reasoning works study models that gather additional information before committing to an answer, often by interacting with tools or environments over multiple turns. LMRL Gym introduces benchmarks for multi-turn reinforcement learning with LLM-based agents, highlighting challenges of long-horizon control and partial observability~\citep{abdulhai2023lmrl}. In high-stakes domains, MedIQ evaluates question-asking LLMs that propose targeted follow-up queries to reduce diagnostic uncertainty in clinical reasoning~\citep{li2024mediq}. Zhou et al.\ move from passive to active reasoning and ask whether LLMs can pose the right questions under incomplete information, proposing benchmarks and methods for query selection when key evidence is missing~\citep{zhou2025passive}. \method{} instantiates a similar active reasoning loop in the visual domain: a text-only reasoner must decide which perception queries to issue to a noisy VLM sensor, interpret self-consistency as a coarse uncertainty signal, and determine when the accumulated evidence is sufficient to commit to an answer. Unlike standard self-reflection methods~\citep{renze2024self}, which focus on an LLM revising its own internal textual logic, \method{} forces the agent to handle explicitly stochastic, external sensory feedback. To the best of our knowledge, we are among the first to explicitly model uncertainty in the environmental feedback itself within an active reasoning benchmark; moreover, our construction is fully automatic and inexpensive, and does not require hand-crafted text simulators or environment-specific infrastructure.

\section{\method{} Construction}

\subsection{Problem Setup: Active Reasoning with a Noisy Visual Sensor}

We formulate VQA as an active reasoning problem for a text-only LLM reasoner $R$
interacting with a noisy visual sensor $S$. Each instance consists of an image $x$, an original VQA
question $q^{(0)}$, and a ground-truth answer $y$. The sensor $S$ is \emph{stateless}: every call
$S(x, q)$ depends only on $(x, q)$. At interaction step $t$, the reasoner observes the interaction history
\begin{equation}
\begin{split}
    h^{(t)} = \big(&q^{(0)},\,(q^{(1)}, a^{(1)}, u^{(1)}),\,\dots,\\
    &(q^{(t-1)}, a^{(t-1)}, u^{(t-1)})\big),
\end{split}
\end{equation}
and chooses either to conclude with an answer or to issue a self-contained query $q^{(t)}$ to the sensor. To model noisy perception and expose uncertainty, we sample the sensor $K$ times with the same prompt:
\begin{equation}
    \{a^{(t)}_1, \dots, a^{(t)}_K\} \sim S(x, q^{(t)}).
\end{equation} We then (i) sample one reply $a^{(t)}$ uniformly from $\{a^{(t)}_i\}_{i=1}^K$
as the realized sensory feedback, and (ii) estimate an uncertainty score by computing the empirical
consistency of this answer:
\begin{equation}
    u^{(t)} = \sum_{i=1}^{K} \mathbf{1}\big[\text{sem\_eq}(a^{(t)}_i, a^{(t)})\big],
\end{equation}
where $\text{sem\_eq}(\cdot,\cdot)$ is implemented by an LLM-based aggregator that judges whether two responses are semantically equivalent. The pair $(a^{(t)}, u^{(t)})$ is appended to the history
to form $h^{(t+1)}$. The interaction loop continues until $R$ outputs a parsable answer or the step
budget $T$ is reached.

\subsection{Dataset Creation}
\noindent \textbf{Dataset Sources}. We construct our benchmark from multiple-choice VQA datasets that collectively cover diverse visual reasoning abilities. Concretely, we include three broad categories: perception on natural images, using SeedBench \citep{li2023seed} for everyday scene understanding; chart understanding, using FigureQA \citep{kahou2017figureqa}; and knowledge-based VQA, using MMStar \citep{chen2024we}, CVQA \citep{romero2024cvqa}, and ReasonVQA \cite{tran2025reasonvqa} to capture reasoning settings that require discipline-specific, cultural, and encyclopedic knowledge.

\begin{figure*}[t]
\centering
\includegraphics[width=0.95\textwidth, trim={0cm 0.1cm 0cm 0cm}, clip]{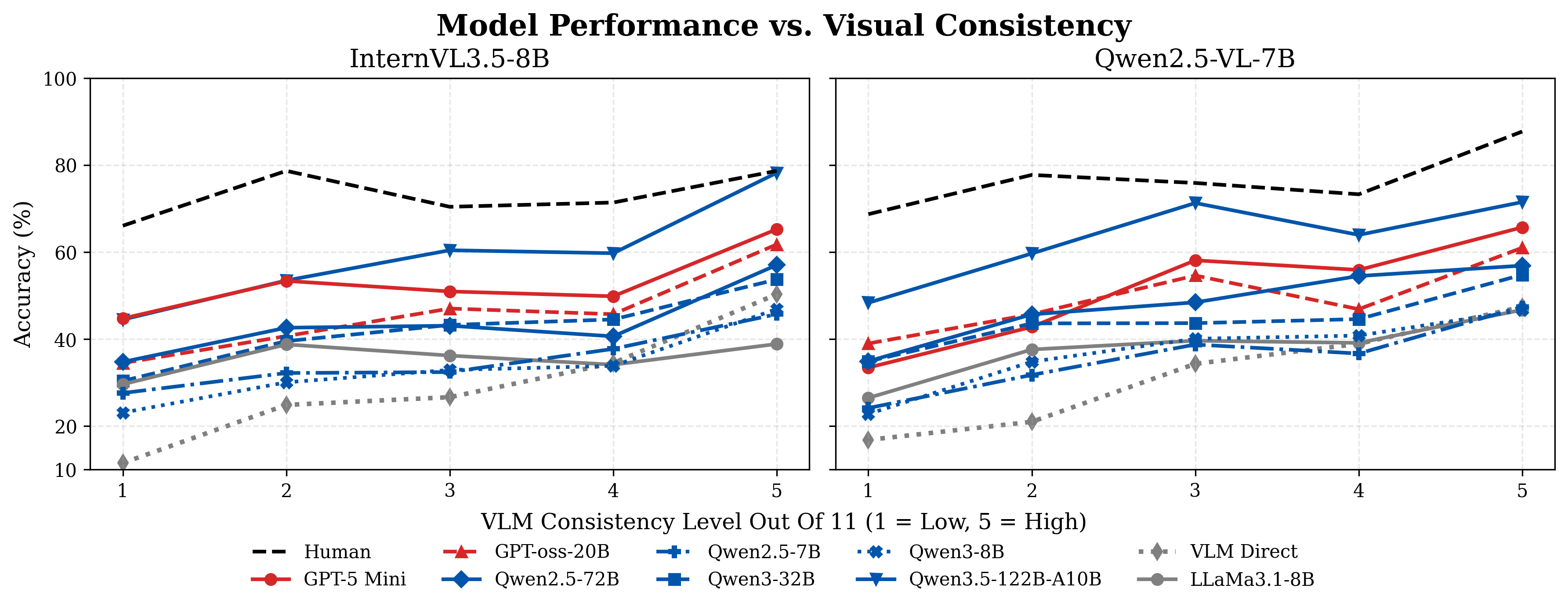}
\vspace{0.01cm}
\caption{Model performance using BP vs.\ VLM consistency level. The x-axis shows binned levels of how often the VLM gives the correct answer (out of 11 samples), with 1 being the lowest consistency and 5 being the highest. LLM reasoners improve over the raw VLM baseline (gray), with humans (dashed black) maintaining the highest performance across all consistency levels.}
\vspace{-0.5cm}
\label{fig:combined_onsistency}
\end{figure*}

\noindent \textbf{VLM Sensors}. We deliberately utilize two modern VLM sensors, Qwen2.5-VL-7B and InternVL3.5-8B, chosen because preliminary experiments indicate that they exhibit complementary behaviors and failure modes. We report the VLM behavior results in App.~\ref{sec: correlation_accuracy_consistency} and App.~\ref{sec: ece}. Since medium-sized VLMs are not perfectly calibrated, they may yield inconsistent outputs, serving as ideal noisy sensors. Specifically, as our goal is to assess LLMs’ capability, rather than VLMs’ perceptual accuracy, these VLMs better reveal LLMs' weaknesses, where reasoners cannot rely solely on VLM confidence and reasoning abilities to make decisions.

\noindent \textbf{Two-stage Filtering}. To ensure that the reasoning tasks are both challenging and solvable, we adopt a two-stage filtering strategy. First, we apply a consistency-based filter to isolate questions on which the sensors themselves struggle: for each question and sensor, we sample 11 independent responses, compute a consistency score $c$ as the number of times the ground-truth option is predicted, and retain only questions with $1 \le c \le 5$. We use self-consistency by design: it is an observable, reproducible, and externally verifiable signal that a text-only reasoner can directly perceive and act on, and it is a well-established, practical proxy for predictive uncertainty in the literature~\citep{wang2023selfconsistency, kuhn2023semantic, xia2025survey}. This guarantees that the VLM is occasionally correct but overall unreliable, so the reasoner cannot succeed by simply forwarding the original question and must engage in additional information seeking. Second, human annotators assume the role of a blind LLM reasoner: they interact with the VLM sensor via text-only queries (without seeing the image), and we keep only questions that are correctly solved by at least one annotator. Finally, we obtain 100 high-quality questions per (sensor, dataset) pair, yielding 1,000 challenging reasoning instances spanning a diverse range of VQA skills and VLM failure modes. This two-stage filter serves as a quality-control criterion: it discards instances where the VLM is either reliable or unsolvable even in theory, retaining only instances where the noisy evidence is logically resolvable by careful reasoning.

\section{Experiments and Analysis}
\subsection{Baselines and Metrics}
We evaluate a range of LLM reasoners spanning proprietary and open-source families: GPT-5 Mini, GPT-oss-20B, Qwen3.5-122B-A10B, Qwen3 (8B, 32B), Qwen2.5 (7B, 72B), and LLaMa3.1-8B. For each model, we consider three reasoning strategies. (i) \emph{ReAct} \citep{yao2023react}, where the model alternates between textual thoughts and actions, and actions correspond to querying the VLM sensor. (ii) \emph{Uncertainty of Thoughts (UoT)} \citep{hu2024uncertainty}, adapted from the original information-seeking setting to VQA by modifying the prompt to describe our setup instead of medical diagnosis. (iii) \emph{Belief-state prompting (BP)}, a prompting recipe we design around the failure modes we hypothesize models exhibit under noisy evidence, which we later substantiate in Sections~\ref{sec:belief_analysis} and~\ref{sec:query_strategy}. BP instructs the model to track its internal belief over answer options during the interaction and to commit to a final choice only when sufficiently certain. While we do not claim BP as a core contribution, it serves as a strong task-specific prompting ceiling to ensure underperformance is not caused by inadequate instructions. 

We report \emph{accuracy} and \emph{Expected Calibration Error} (ECE), computed from cross-run prediction consistency, as our primary metrics~\citep{kostumov2024uncertaintyaware}. For non-proprietary models, we draw 5 trajectories per instance at temperature 1.0 and compute ECE using the agreement of the predicted option across runs. For GPT-5 Mini, due to cost constraints, we obtain a single response per instance.
 
In addition, we collect a human baseline by randomly sampling 50 questions for each setting and asking a different human annotator to solve them, using the same interaction protocol. In Perception (SeedBench) and Charts (FigureQA), human annotators had no access to external tools, yet the human--model gap persists, suggesting it reflects reasoning and uncertainty-handling deficits rather than information access. For knowledge-heavy datasets (MMStar, CVQA, ReasonVQA), annotators were allowed limited use of search engines or AI tools only to clarify unfamiliar terms or background knowledge required to understand the question, but not to carry out the main reasoning or come up with the final answer. This keeps the comparison fair: the benchmark targets uncertainty handling rather than penalizing annotators for lacking high-level domain knowledge in novel datasets. See App.~\ref{sec: knowledge breakdown} for detailed per-dataset results.

\noindent \textbf{LLM use disclosure.} We use LLM-based judges for semantic grouping in Eq. (3) and for query-type annotation in Section~\ref{sec:query_strategy}; details and validation are provided in the appendix.
 
\begin{figure}[t]
  \centering
  \includegraphics[width=\columnwidth]{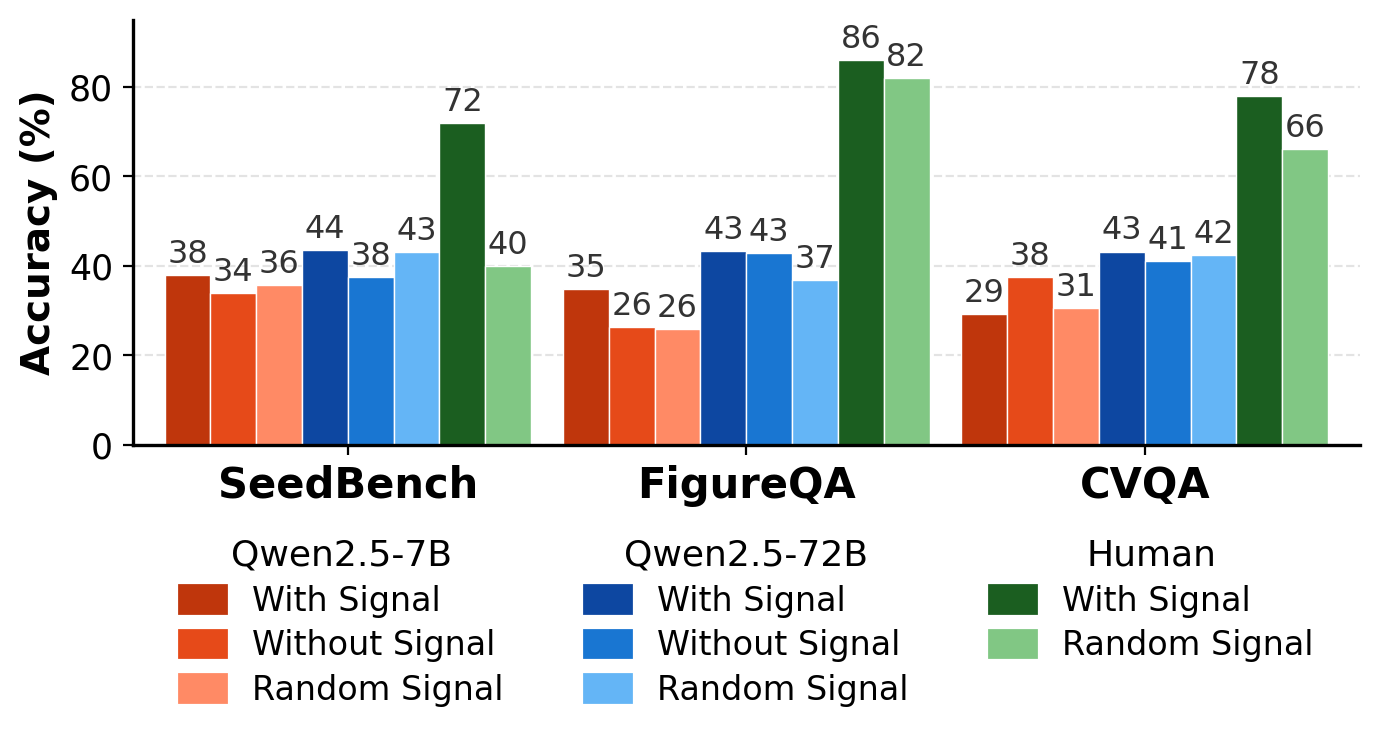}
  \caption{Ablation study evaluating the impact of the consistency signal on humans and models under the Belief-State Prompting (BP) strategy.}
  \label{fig:ablation_study}
\end{figure}
\subsection{Analysis and Observations}
\noindent \textbf{Tasks are genuinely hard even for strong LLMs}. The strongest models---Qwen3.5-122B-A10B, GPT-5 Mini, and Qwen2.5-72B---achieve only 40--60\% accuracy in most category-level settings (at best 67.7\%), while human performance reaches roughly 70--90\%, underscoring that robustly reasoning under noisy visual evidence remains an open challenge for current LLMs. Among LLM reasoners, Qwen3.5-122B-A10B leads overall, followed by GPT-5 Mini; within each family, larger models generally outperform their smaller counterparts, confirming that scale partially mitigates---but does not close---the gap to human performance. Results are qualitatively consistent across the two VLM sensors, though performance is commonly higher with Qwen2.5-VL-7B. 

Among strategies, BP and ReAct are the two strongest, while UoT trails in accuracy. BP, which we treat as a task-specialized baseline rather than a methodological contribution, helps most for the strongest reasoner: Qwen3.5-122B-A10B. It increases accuracy over ReAct in most settings and achieves among the lowest ECE. For weaker models, ReAct is usually on par or better, consistent with belief tracking paying off mainly when the reasoner follows it faithfully (Sec.~\ref{sec:belief_analysis}).

\noindent \textbf{LLM reasoners improve over raw VLM baselines}. On this explicitly filtered pool, inference-only LLM reasoners generally outperform direct VLM predictions, yielding higher accuracy and lower ECE. For instance, with the Qwen2.5-VL-7B sensor, direct VLM achieves a highly miscalibrated ECE of 0.49 on Perception tasks, while querying it via Qwen3.5-122B-A10B (BP) reduces the ECE to 0.22 and more than doubles the accuracy. This suggests that strong LLMs can act as ``cognitive error-correctors'' for noisy perceptual systems, providing the largest lift over raw VLM baseline precisely when the VLM's internal consistency is lowest.

\noindent \textbf{Effect of VLM consistency}. Figure~\ref{fig:combined_onsistency} shows the accuracy of humans, VLM-direct, and LLM-based reasoners as a function of the VLM self-consistency level. Across both sensors, LLM performance increases with consistency; however, the gap between VLM-direct and LLM-based systems narrows in the easiest bins, and only the largest reasoners achieve substantial gains over the VLM baseline there. In contrast, human blind performance degrades only mildly in the lowest-consistency bins, indicating a large performance gap in the most difficult settings. In addition, reasoners paired with the better-calibrated Qwen2.5-VL benefit more from increasing consistency than those using InternVL3.5, while human accuracy remains relatively stable even with miscalibrated signals.

\begin{table}[t]
\centering
\small
\setlength{\tabcolsep}{3.5pt}
\resizebox{\columnwidth}{!}{%
\begin{tabular}{l ccccc}
\toprule
 & SeedBench$^\dagger$ & FigureQA & CVQA & MMStar & ReasonVQA \\
\midrule
\multicolumn{6}{l}{\textit{InternVL3.5-8B sensor}} \\
Qwen3-8B & 38.8 & 28.6 & 26.0 & 32.0 & 38.2 \\
\quad+\,SFT & 36.0 & \textbf{30.6} & \textbf{27.6} & \textbf{34.0} & 26.8 \\
\midrule
\multicolumn{6}{l}{\textit{Qwen2.5-VL-7B sensor}} \\
Qwen3-8B & 38.4 & 30.8 & 29.0 & 38.4 & 48.2 \\
\quad+\,SFT & \textbf{45.8} & \textbf{32.6} & 20.6 & 32.0 & 46.8 \\
\bottomrule
\end{tabular}%
}
\caption{BP accuracy (\%) before and after self-distillation on 1,800 successful SeedBench trajectories. $^\dagger$In-distribution.}
\label{tab:self_distill}
\end{table}

\noindent \textbf{Ablation study of the consistency signal.} Figure~\ref{fig:ablation_study} compares BP performance of Qwen2.5-7B/72B with the true VLM answer-consistency metadata, without the metadata, and with a randomly injected signal. We observe sizable gains on most perception and diagram datasets when using the true signal, indicating that reasoners can exploit this uncertainty cue rather than just the presence of numerical metadata. In contrast, the signal can sometimes hurt on the knowledge-based benchmark (CVQA): the VLM can sometimes be confidently wrong, making true consistency a noisy or even adversarial proxy for reliability (see App.~\ref{sec:error_analysis} for a qualitative example). By contrast, human accuracy drops sharply under a random signal on SeedBench and CVQA, suggesting that humans rely on the signal more than LLMs do. This presents a challenge as a reasoning task and suggests a direction for future work where reasoners learn domain-specific strategies.

\subsection{Can \method{} be solved with basic fine-tuning?}
\label{sec:self_distill}

As a first step, we explore whether \method{} adaptive behavior can be learned through standard supervised fine-tuning (SFT), instead of relying only on inference-time prompting. We test a self-distillation approach, fine-tuning a representative model, Qwen3-8B, on its own successful belief-state prompting (BP) traces.

\noindent \textbf{Setup.}  We collect 1,800 successful BP trajectories from a held-out SeedBench split, balanced across InternVL3.5-8B and Qwen2.5-VL-7B. We apply LoRA fine-tuning (rank 16, $\alpha=32$) for 7 epochs, optimizing cross-entropy on the assistant's turns, and select the checkpoint by best validation loss.

\noindent \textbf{Results.}
As shown in Table~\ref{tab:self_distill}, self-distillation does not yield consistent gains: accuracy improves on some combinations (in-distribution SeedBench with Qwen2.5-VL-7B, FigureQA with both sensors) but degrades on others (CVQA with Qwen2.5-VL-7B, ReasonVQA with InternVL3.5-8B), with improvements remaining sensor- and dataset-specific. This suggests that the model memorizes sensor-specific patterns instead of the adaptive strategy \method{} requires, formulating queries based on prior evidence and switching tactics when direct inference fails. This aligns with recent findings that supervised fine-tuning tends to fit surface patterns~\citep{chu2025sft, guo2025deepseekr1}. We thus read this as a preliminary probe: within our setup, cloning successful trajectories does not internalize active reasoning, and the fine-tuned model remains far below humans (Table~\ref{tab:category-results}), underscoring the difficulty of \method{} and leaving it as an open training challenge. 

\subsection{Retrospective Belief-State Analysis}
\label{sec:belief_analysis}

To understand how models make decisions (whether they terminate prematurely or update their belief states faithfully) and to explain the performance gap between Qwen3.5-122B-A10B and Qwen2.5-7B, we retrospectively elicit their implicit belief states during reasoning. While this reflects the model's self-reported interpretation rather than guaranteed internal states, the elicitation reveals consistent differences between the two reasoners.

\noindent \textbf{Methodology.} For each step $t$ of a BP conversation, we replay the trajectory (2000 trajectories in total over 4 settings) and prompt the original LLM to output its implicit belief state as a probability distribution $b_t \in \Delta^{|\mathcal{O}|}$ over the answer options (details of our setting in App.~\ref{sec:belief_probe_prompt}). We track belief convergence via $p_{\mathrm{correct}}(t) = b_t[\text{ground-truth option}]$ and final confidence via $p_{\mathrm{chosen}}$ (the probability assigned to the ultimately selected option).

\begin{figure}[t]
    \centering
    \includegraphics[width=\linewidth]{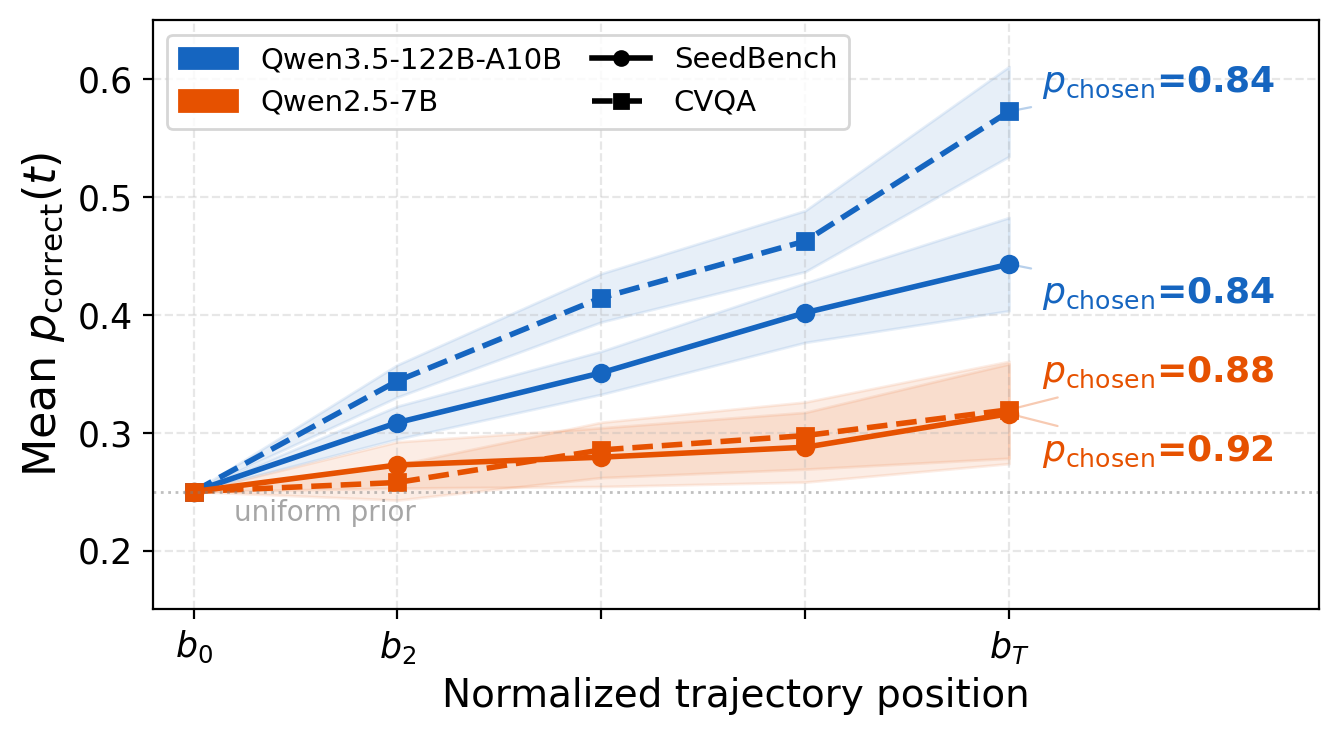}
    \caption{Mean $p_\mathrm{correct}(t)$ over the belief trajectory ($b_0 \to b_T$). The figure also shows $p_\mathrm{chosen}$ (confidence in the chosen answer) at termination.}
    \label{fig:belief_trajectories}
\end{figure}

\noindent \textbf{Reported Confidence vs. Belief Convergence.}
Tracking $p_{\mathrm{correct}}$ over time (Figure~\ref{fig:belief_trajectories}) reveals differences in how the two models update beliefs. Qwen3.5-122B-A10B's trajectories are consistent with evidence-driven convergence: its $p_{\mathrm{correct}}$ rises steadily from the uniform prior to 0.44 (SeedBench) and 0.57 (CVQA) at termination, while Qwen2.5-7B's $p_{\mathrm{correct}}$ rises only marginally (from 0.25 to about 0.32). Despite this limited convergence, Qwen2.5-7B shows what we call \emph{performative confidence} at termination, reporting a higher mean $p_{\mathrm{chosen}}$ than the 122B model. Moreover, Qwen2.5-7B's wrong answers carry even higher confidence (0.93) than its correct ones (0.89), suggesting its stated confidence follows the prompt's 90\% threshold more closely than the strength of the gathered evidence.

\noindent \textbf{Premature Commitment and Early Anchoring.}
One failure mode we observe for the smaller model is deciding too early. Qwen2.5-7B frequently anchors prematurely: 46\% of its SeedBench trajectories already exceed 0.7 confidence after the first evidence step, compared to 13\% for Qwen3.5-122B-A10B. As shown in Table~\ref{tab:convergence_speed} (App.~\ref{rounds_number}), Qwen2.5-7B reaches high confidence roughly 1.6$\times$ faster than the 122B model. Notably, its \emph{incorrect} trajectories commit even faster than the 122B model's \emph{correct} ones (1.9 vs.\ 3.1 steps on SeedBench; 2.6 vs.\ 3.6 on CVQA), suggesting that the smaller model's speed reflects early anchoring rather than stronger evidence. In contrast, the 122B model gathers more evidence before committing, consistent with its steadier rise in $p_{\mathrm{correct}}$. 

Within our two-model comparison, these results suggest that faithful belief tracking under uncertainty depends on reasoner capability rather than being induced by the prompt alone.

\subsection{Querying Strategy}
\label{sec:query_strategy}
A key question is \emph{how} human annotators achieve their large accuracy advantage. We classify each of 1,937 queries from human and model performances on SeedBench and CVQA into five types---\textit{Descriptive}, \textit{Targeted}, \textit{Contrastive}, \textit{Confirmatory}, and \textit{Comparative}---using an LLM judge (the categories are defined in App.~\ref{sec:query_classify}). Table~\ref{tab:query_strategy} reports the distribution of the three most frequent query types across models and humans split by round. 

\begin{table}[t]
\centering
\setlength{\tabcolsep}{4.5pt}
\resizebox{\columnwidth}{!}{%
\begin{tabular}{l cc ccc}
\toprule
& \multicolumn{2}{c}{\textbf{Round 1}} & \multicolumn{3}{c}{\textbf{Round 2+}} \\
\cmidrule(lr){2-3} \cmidrule(lr){4-6}
 & Descriptive & Targeted & Descriptive & Targeted & Confirmatory \\
\midrule
Human                & \textbf{97\%} & 3\%  & 26\% & 56\% & 13\% \\
Qwen3.5-122B-A10B    & 75\%          & 25\% &  8\% & 68\% & 15\% \\
Qwen2.5-7B           & 38\%          & 57\% &  2\% & 60\% & \textbf{23\%} \\
\bottomrule
\end{tabular}%
}
\caption{Distribution of the three most common query types by round (averaged over SeedBench and CVQA).}
\label{tab:query_strategy}
\end{table}
\noindent\textbf{Opening strategy.}
Humans begin with broad scene-overview queries in virtually every conversation (97\% descriptive in Round 1), building a global mental model before narrowing to specific evidence. Qwen3.5-122B-A10B follows a similar but weaker pattern (75\%), while Qwen2.5-7B skips the initial scene survey in 62\% of conversations, jumping directly to narrow targeted questions without first establishing global context.

\noindent\textbf{Confirmatory bias in weaker models.}
In subsequent rounds, the sharpest divergence is the \textit{confirmatory} rate: Qwen2.5-7B seeks VLM agreement (``Is it true that\ldots'', ``So the answer is (B)?'') 23\% of the time, nearly double the human rate (13\%). Rather than stress-testing evidence by challenging the VLM's earlier answer, the weaker model tends to solicit validation of its initial hypothesis. This overconfident acceptance of early evidence is consistent with the premature commitment pattern described in Section~\ref{sec:belief_analysis}. The gradient in confirmatory rate (Qwen2.5-7B $>$ Qwen3.5-122B-A10B $>$ Human) tracks model capability inversely, suggesting that stronger reasoners more often adopt a deliberate explore-then-verify strategy.

Overall, the annotators often outperform the models through a more deliberate querying strategy. They first request a broad scene description to build an initial mental model of the image, then narrow to question-relevant details. Importantly, rather than trusting a single answer, they also issue contrastive follow-up questions to verify whether the VLM’s responses remain consistent under rephrasing. 

\subsection{Additional Analysis and Error Attribution}
\label{sec:additional_analysis_main}

\noindent We validate our automated semantic equivalence aggregator against expert human annotators in App.~\ref{sec:additional_analysis}. The LLM judge demonstrates high agreement, confirming its reliability as a scalable proxy for determining empirical consistency. Furthermore, we explore the trade-off between computational cost and the consistency signal by increasing the VLM sampling budget from $K=5$ to $K=11$, which slightly lowers ECE and Standard Deviation, while the raw accuracy remains similar.

To contextualize the performance gaps and querying behaviors, App.~\ref{sec:error_analysis} details two prominent failure modes. First, in knowledge-intensive domains (e.g., CVQA), the VLM sensor can be \emph{confidently wrong}. In these cases, empirical consistency becomes a trap, and the reasoner must rely heavily on reasoning over descriptive features rather than VLM assertions. Second, when confronted with self-contradictory evidence, models frequently exhibit premature commitment. Unlike annotators who dynamically adjust their strategies based on the provided signal, LLM reasoners often stubbornly accept the highest-confidence response.

\section{Conclusion}
Real-world reasoning depends not just on gathering more evidence, but on deciding what to trust when the evidence itself is noisy. \method{} makes this challenge measurable by casting VQA as active reasoning through unreliable visual sensors: a reproducible 1,000-instance benchmark of cases where the sensor is inconsistent yet the problem remains solvable by humans, isolating uncertainty handling from raw perception and providing complementary failure modes. Even strong LLMs benefit from consistency signals but remain well below human performance, calling for methods that can actively probe, reconcile, and search over conflicting evidence.

\section*{Limitations}

\noindent Our study focuses on a controlled source of unreliable evidence: the stochastic perceptual noise from a visual sensor. This scope is our intentional design choice, as it lets us study reasoning under uncertainty separately from perception itself. However, we do not claim to characterize or measure the full space of noise and unreliability that reasoners face in practice, and it can arise from other sources, such as retrieval and web search, interactions with external tools, or modalities beyond vision. We therefore view our setting as one representative instance of a broader phenomenon, and we hope that a similar style of analysis can be extended to these additional sources of unreliability in future work.

\bibliography{custom}

\appendix
\onecolumn

\section{Additional Analysis}
\label{sec:additional_analysis}

\subsection{Frontier VLM Sensor Pilot}
To probe how the benchmark behaves when the visual sensor is a frontier model, we ran a small pilot study using Gemini 3.1 Flash Lite as the sensor with a Qwen3.5-122B-A10B reasoner on 100 InternVL-uncertain instances per dataset (and Gemini 2.5 Flash on FigureQA), following the same interaction protocol as the main table. Although a rigorous two-stage filtering would extract frontier-VLM-uncertain instances separately, which is too costly and resource-intensive, we treat this as an insight into the frontier-sensor regime. As shown in Table~\ref{tab:gemini_pilot_app}, when Gemini is generally reliable---as in most settings---there is no substantial difference between the reasoner under the interactive setting and VLM Direct; indeed, using an LLM reasoner can sometimes hurt the VLM Direct baseline. When Gemini 3.1 Flash Lite is genuinely uncertain (FigureQA: 37.8\%), Qwen3.5-122B-A10B improves the VLM Direct baseline by around 30.8 percentage points, even exceeding the InternVL-based gain. Even though this pilot uses InternVL-uncertain instances, frontier VLMs are less suitable as sensors in our setting: as the sensor becomes reliable, the task reduces to trusting it and the reasoning challenge for the LLM degrades, whereas medium-scale VLMs expose the noisy-evidence regime the benchmark targets.

\begin{table}[h]
\centering
\begin{tabular}{lccc}
\toprule
\textbf{Dataset} & \textbf{Gemini VLM Direct} & \textbf{BP (Gemini sensor)} & \textbf{Paper BP (InternVL sensor)} \\
\midrule
CVQA      & \textbf{85.2\%} & 81.0\% & 61.2\% \\
FigureQA  & 37.8\% & \textbf{68.6\%} & 59.8\% \\
MMStar    & 62.1\% & \textbf{67.0\%} & 50.6\% \\
ReasonVQA & \textbf{76.1\%} & 74.6\% & 71.2\% \\
SeedBench & 49.1\% & \textbf{50.4\%} & 48.6\% \\
\bottomrule
\end{tabular}
\caption{Frontier VLM sensor pilot: BP accuracy with a Gemini 3.1 Flash Lite sensor (Qwen3.5-122B-A10B reasoner, 100 InternVL-uncertain instances per dataset) compared against the Gemini VLM Direct baseline and the paper's BP with the InternVL3.5-8B sensor. Observing that Gemini 3.1 Flash Lite did not perform well on FigureQA, we tested another evaluation with Gemini 2.5 Flash as a sensor: VLM Direct 61.6\%, BP 67.0\%.}
\label{tab:gemini_pilot_app}
\end{table}

\subsection{LLM Judge Validation}
To validate the automated semantic equivalence judge, two expert annotators independently evaluated 100 sampled response lists (balanced between those judged by Qwen2.5-7B and 72B). For each list of five VLM responses, annotators assigned a confidence score based on semantic grouping. Initial inter-annotator discrepancies were resolved through collaborative discussion to establish a consensus ground truth, which we then used to benchmark the LLM-based judge's performance. We calculate both the agreement score and Cohen's Kappa Metric to compare human judgments against the automated LLM judges, which are shown in Table~\ref{tab:llm_judge_validation_app}. We observe high within-1-point agreement and moderate exact agreement. We also calculated a Quadratic Weighted Cohen's Kappa and a Pearson correlation. These high values demonstrate that our LLM judge is a robust proxy for human semantic grouping. Specifically, both Qwen2.5-7B and 72B have high Pearson correlations hovering around 0.80, suggesting that the LLM judge is consistently calibrated across different model sizes.

\begin{table}[h]
\centering
\begin{tabular}{lc}
\toprule
\textbf{Evaluation Metric} & \textbf{Value} \\
\midrule
Exact Agreement & 66.0\% \\
Within-1 Agreement & 91.0\% \\
Mean Absolute Error (MAE) & 0.49 \\
Cohen's Kappa (Quadratic) & 0.79 \\
Pearson Correlation & 0.80 \\
\bottomrule
\end{tabular}
\caption{Validation of the LLM semantic equivalence judge against human annotators.}
\label{tab:llm_judge_validation_app}
\end{table}

\subsection{Trade-off between computation cost and consistency signal}
We expanded the sampling budget for InternVL3.5-8B to $K = 11$, instead of $K = 5$, on 300 SeedBench, FigureQA, and CVQA instances (100 samples for each dataset) for Qwen2.5-7B and 72B. While raw accuracy remains similar, the Expected Calibration Error (ECE) and Standard Deviation slightly decrease.  Table~\ref{tab:acc_const_tradeoff_app}, which outlines the BP performance and calibration differences, shows a modest trade-off: increasing computational cost (more sensor samples) stabilizes the uncertainty signal.

\begin{table}[h]
\centering
\resizebox{\columnwidth}{!}{
\begin{tabular}{llcccccc}
\toprule
 & & \multicolumn{2}{c}{\textbf{SeedBench}} & \multicolumn{2}{c}{\textbf{FigureQA}} & \multicolumn{2}{c}{\textbf{CVQA}} \\
\cmidrule(lr){3-4} \cmidrule(lr){5-6} \cmidrule(lr){7-8}
\textbf{Reasoner} & \textbf{VLM Samples ($K$)} & \textbf{Acc $\pm$ Std} & \textbf{ECE} & \textbf{Acc $\pm$ Std} & \textbf{ECE} & \textbf{Acc $\pm$ Std} & \textbf{ECE} \\
\midrule
Qwen2.5-72B & $K=5$  & 43.6 $\pm$ 3.51 & 0.31 & 43.4 $\pm$ \textbf{4.28} & 0.41 & 43.2 $\pm$ 2.68 & \textbf{0.19}\\
            & $K=11$ & 44.4 $\pm$ \textbf{1.82} & \textbf{0.27} & 42.6 $\pm$ 4.72 & \textbf{0.40} & 43.4 $\pm$ \textbf{2.30} & 0.20 \\

\midrule

Qwen2.5-7B  & $K=5$  & 38.0 $\pm$ 4.00 & 0.32 & 34.8 $\pm$ 6.46 & \textbf{0.25} & 29.2 $\pm$ 6.14 & 0.32\\
            & $K=11$ & 37.4 $\pm$ \textbf{2.19} & \textbf{0.27} & 31.8 $\pm$ \textbf{1.30} & 0.39 & 32.8 $\pm$ \textbf{2.95}  & \textbf{0.23} \\
\bottomrule
\end{tabular}
}
\caption{Ablation study showing trade-offs between computation cost (number of VLM samples $K$) and the consistency signal on SeedBench, FigureQA, and CVQA.}
\label{tab:acc_const_tradeoff_app}
\end{table}



\newpage
\section{Qualitative Error Analysis}
\label{sec:error_analysis}
We present two failure cases that illustrate qualitatively distinct error modes: one in which the VLM sensor is \emph{confidently wrong}, rendering the consistency signal uninformative, and one in which the VLM provides \emph{self-contradictory} evidence that a human reconciles but models do not. Together, these cases concretize the quantitative gaps reported in the main text.

\subsection{VLM Confidently Wrong on Domain Knowledge}

\noindent\textbf{Question:} \textit{What is the name of this building in Mongolia? \\(A) Hunnu Mall \quad (B) Department Store \quad (C) Ulaanbaatar Department Store \quad (D) Sport Center} \hfill \textbf{GT: (C)}

\smallskip

This CVQA (knowledge-intensive) instance asks a reasoner to identify a Mongolian landmark (see the full interaction trace in App.~\ref{sec: traces}). We highlight the key contradiction. When prompted to confirm the building's identity, the VLM answers:

\smallskip
\small
\begin{tabular}{@{}p{0.06\columnwidth}p{0.85\columnwidth}@{}}
\textbf{R2} & \textit{Q:} Can you confirm whether the building is the Hunnu Mall, Department Store, Ulaanbaatar Department Store, or Sport Center? \\
& \textit{VLM:} The building in the image appears to be the \textbf{Ulaanbaatar Department Store}. \textbf{(4/5)} \\[6pt]
\textbf{R4} & \textit{Q:} Are there any signs, labels, or other indicators in the image that specifically identify the building as the Ulaanbaatar Department Store? \\
& \textit{VLM:} \textbf{No}, the image does not show any visible signage or labels that clearly identify the building as the Ulaanbaatar Department Store. \textbf{(4/5)} \\[3pt]
& \textit{Reasoner:} ``The VLM's response introduces more uncertainty \ldots'' $\to$ \textbf{[revises belief downward]}
\end{tabular}
\normalsize
\smallskip
R2 and R4 are in tension: if the building \emph{is} the Ulaanbaatar Department Store, some perceptual basis must underlie that claim---yet R4 explicitly denies that any such basis is visible. Crucially, both answers carry \emph{identical} 4/5 confidence. The consistency signal therefore provides no discriminative information: the VLM is equally self-assured when asserting and when retracting the same fact. The reasoner is forced into genuine uncertainty, where it has to place more trust in the descriptive information to resolve the problem.

\subsection{Human Inconsistency Detection vs.\ Model Acceptance}


\noindent\textbf{Question (SeedBench):} \textit{What is the relative position of the woman's boots and jeans in the image? (A) The jeans are overlapping the boots. (B) The boots are overlapping the jeans. (C) The boots and jeans are next to each other. (D) The boots and jeans are not visible in the same frame.} \hfill \textbf{GT: (C)}

\smallskip

This SeedBench instance demonstrates the contrasting strategies of a human annotator and two LLM reasoners when confronted with self-contradictory VLM feedback.

\smallskip
\noindent\textbf{Human annotator} (9 rounds, correct). The annotator begins with a broad scene description (u{=}2/5), then isolates each item individually before asking the VLM directly about overlap:

\small
\begin{tabular}{@{}p{0.06\columnwidth}p{0.85\columnwidth}@{}}
\textbf{R5} & \textit{Q:} What is the relative position between the jeans and the boots? \\
& \textit{VLM:} The jeans are worn over the boots. \textbf{(4/5)} \\[3pt]
\textbf{R6} & \textit{Q:} What is the relative position between the boots and the jeans? \\
& \textit{VLM:} The boots are worn over the jeans. \textbf{(5/5)}
\end{tabular}
\normalsize
\smallskip

\noindent The VLM gives logically contradictory answers in consecutive rounds: ``jeans over boots'' (4/5) followed by ``boots over jeans'' (5/5). Rather than accepting the higher-confidence answer, the human \emph{changes strategy entirely}, probing what objects are \emph{near} each item:

\small
\begin{tabular}{@{}p{0.06\columnwidth}p{0.85\columnwidth}@{}}
\textbf{R7} & \textit{Q:} What can be found nearby the boots? \\
& \textit{VLM:} \ldots a black handbag and a person wearing \ldots denim shorts\ldots \textbf{(3/5)} \\[3pt]
\textbf{R8} & \textit{Q:} What can be found nearby the jeans? \\
& \textit{VLM:} \ldots a black beret, a gray knitted sweater, black boots\ldots \textbf{(3/5)} \\[3pt]
& $\to$ \textbf{(C) Next to each other} \ding{51}
\end{tabular}
\normalsize
\smallskip

\noindent By triangulating from a different angle, the annotator infers that the items sit side-by-side rather than overlapping---a conclusion that neither direct-overlap answer supported.

\smallskip
\noindent\textbf{Qwen2.5-7B} (4 rounds, wrong). The model receives the same contradictory signal---round~1 says ``boots on top'' (2/5) while round~3 says ``jeans overlap boots'' (5/5). Crucially, the model \emph{acknowledges} the contradiction in its own chain-of-thought yet still commits to (A) based on the higher-confidence response.

\smallskip
\noindent\textbf{Qwen3.5-122B-A10B} (5 rounds, wrong). The stronger model fares no better. It detects that rounds~1 and~2 disagree (3/5 vs.\ 1/5), asks two additional clarifying questions, and follows the trend of later, higher-confidence answers toward (B), which is the opposite wrong direction from the 7B model.

\smallskip
\noindent\textbf{Takeaway.} The human's advantage is not simply asking more questions but \emph{switching inference strategy} when evidence is self-contradictory---probing for indirect evidence instead of repeatedly asking the same direct question. Both models detect the contradiction but default to recency or confidence weighting rather than seeking a fundamentally different line of evidence. This qualitative gap aligns with the quantitative querying-strategy analysis (Section \ref{sec:query_strategy}): models issue confirmatory follow-ups where humans pivot to novel angles.

\section{Number of Queries per Setting}
\label{rounds_number}
\noindent \textbf{Number of Queries and Reasoning Strategy.} Figure~\ref{fig:query_counts} reveals distinct trade-offs between reasoning strategies across datasets. Uncertainty-aware prompting (BP and UoT) encourages active information seeking compared to ReAct, which consistently terminates early across all models. However, UoT query counts scale aggressively and inefficiently with model capability, reaching up to 9.8 queries for GPT-5-mini on SeedBench. In contrast, Belief-State Prompting (BP) strikes a calibrated balance. BP consistently drives deeper exploration than ReAct while remaining significantly more efficient than UoT, averaging between 3.6 and 5.3 queries across datasets—a range that most closely mirrors the human strategy (4.8 to 6.4 queries). This targeted querying behavior correlates with the accuracy gains observed in the main paper, consistent with deliberate uncertainty tracking supporting effective interaction. \\ 

\noindent\includegraphics[width=\textwidth]{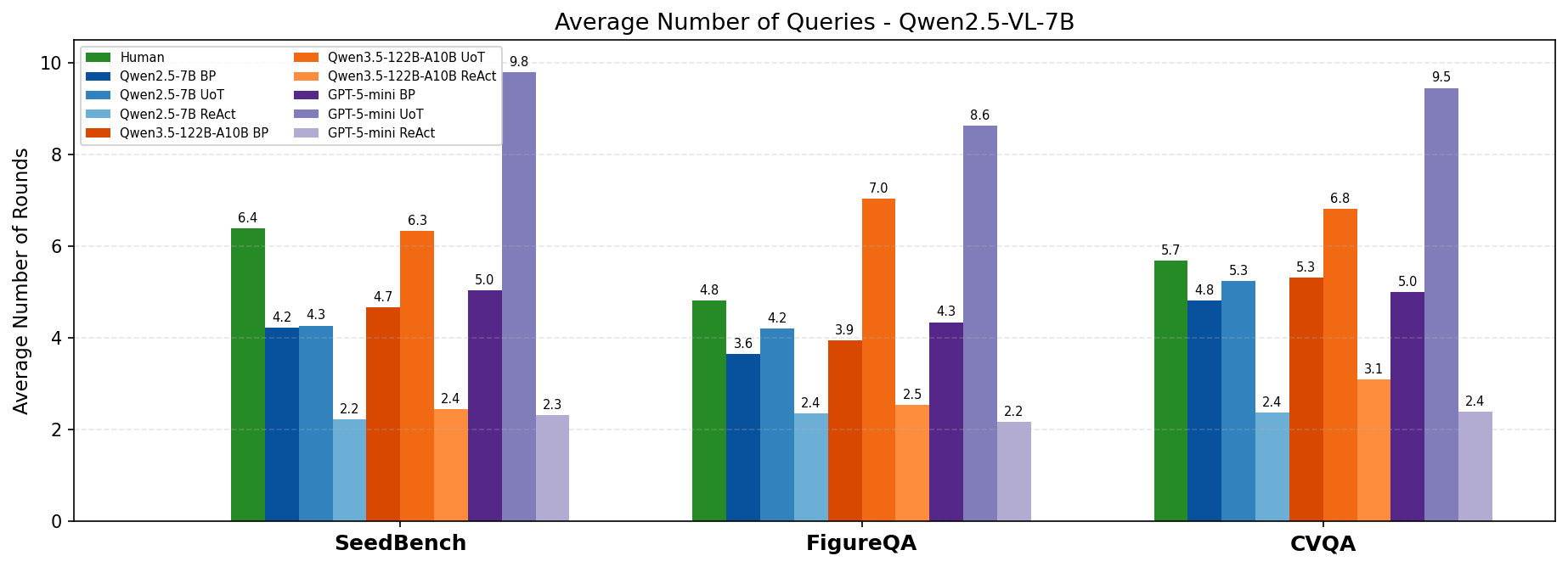}
\captionof{figure}{Average Number of Rounds for 3 Datasets Under 3 Different LLM Baselines and Human Performance.}
\label{fig:query_counts}

\noindent
\\
In our evaluation pipeline, the active reasoning loop terminates under one of three conditions: the model reaches sufficient confidence to select a final answer, exhausts its predefined query budget, or exceeds its context window. In reality, we set a very high maximum round limit $T$ of 100 for both humans and LLMs across all settings, and this does not limit the LLM and human behavior. We have kept track of the conversations and found that there is no question where LLMs failed to conclude due to cutoffs by reaching the maximum round or their context windows. To make sure that humans did not brute-force the VLM sensors, we also report that the average number of queries across all 10 settings is 5.73. \\

\begingroup
\centering
\small
\setlength{\tabcolsep}{4pt}
\begin{tabular}{llcccc}
\toprule
& & \multicolumn{2}{c}{\textbf{Steps to $\max p \ge 0.7$}} & \multicolumn{2}{c}{\textbf{Reach rate}} \\
\cmidrule(lr){3-4} \cmidrule(lr){5-6}
& & Correct & Wrong & Correct & Wrong \\
\midrule
\multirow{2}{*}{SeedBench} & Qwen3.5-122B-A10B & 3.1 & 3.0 & 90\% & 94\% \\
& Qwen2.5-7B & \textbf{1.8} & \textbf{1.9} & 94\% & 98\% \\
\midrule
\multirow{2}{*}{CVQA} & Qwen3.5-122B-A10B & 3.6 & 4.1 & 92\% & 88\% \\
& Qwen2.5-7B & \textbf{2.1} & \textbf{2.6} & 97\% & 92\% \\
\bottomrule
\end{tabular}
\captionof{table}{Convergence speed: mean steps to reach $\max_i\, b_t[i] \ge 0.7$, split by final-answer correctness. Qwen2.5-7B reaches high confidence about 1.6$\times$ faster than the 122B model on both datasets, and even its incorrect trajectories commit earlier than the 122B model's correct ones.}

\label{tab:convergence_speed}
\par
\endgroup

\section{Compute and Latency Cost}
\label{sec:compute_cost}

\noindent \textbf{Wall-clock latency.} We report the end-to-end latency of the three interaction strategies. We measure the average latency (seconds per instance) of VLM Direct, ReAct, and BP for both visual sensors on 100 SeedBench instances (batch size 16, single trial), where the text reasoner (Qwen2.5-7B) and both sensors (InternVL3.5-8B and Qwen2.5-VL-7B) are served locally with vLLM on a single H200 GPU. A single VLM query averages 0.19\,s for InternVL3.5-8B and 0.25\,s for Qwen2.5-VL-7B. As reported in Table~\ref{tab:latency_app}, the interactive strategies add a modest reasoning overhead on top of the sensor: ReAct averages 1.05--1.07\,s per instance and BP averages 2.42--2.48\,s per instance, against 0.19--0.25\,s for VLM Direct.

\begin{table}[h]
\centering
\begin{tabular}{lcc}
\toprule
\textbf{Method} & \textbf{InternVL sensor} & \textbf{Qwen2.5-VL sensor} \\
\midrule
VLM Direct (no text model) & 0.19\,s & 0.25\,s \\
ReAct & 1.05\,s & 1.07\,s \\
BP & 2.42\,s & 2.48\,s \\
\bottomrule
\end{tabular}
\caption{Average wall-clock latency (seconds per instance) of VLM Direct, ReAct, and BP for both visual sensors, measured on 100 SeedBench instances. The text reasoner (Qwen2.5-7B) and both sensors are served locally with vLLM on a single H200 GPU.}
\label{tab:latency_app}
\end{table}

\noindent \textbf{Number of VLM calls.} The number of VLM inference calls scales linearly with the number of interaction rounds: a reasoner issuing $r$ rounds makes $K \times (r-1)$ VLM calls, since each perception query draws $K=5$ samples and the final round is an answer commit that issues no query, while VLM Direct makes a single call per question. Table~\ref{tab:vlm_calls_app} reports the average per dataset. BP is the most query-intensive among the strategies in Table~\ref{tab:vlm_calls_app} (14.7--26.8 calls per question) and ReAct the least among the interactive strategies (7.0--14.5), reflecting BP's deeper information seeking. Aggregated over a full run, the benchmark remains inexpensive to reproduce: solving 1{,}000 instances with the most verbose reasoner (BP with Qwen3.5-122B-A10B) issues on average 3{,}961 perception queries, or roughly 19{,}800 VLM inference calls at $K=5$; at approximately 0.2\,s per call, the total VLM inference time is about 66 minutes on a single H200 GPU with vLLM.

\begin{table}[h]
\centering
\resizebox{\columnwidth}{!}{
\begin{tabular}{lccccc}
\toprule
\textbf{Dataset} & \textbf{VLM Direct} & \textbf{BP (InternVL)} & \textbf{BP (Qwen2.5-VL)} & \textbf{ReAct (InternVL)} & \textbf{ReAct (Qwen2.5-VL)} \\
\midrule
SeedBench & 1 & 18.4 & 18.4 & 7.0 & 7.2 \\
FigureQA & 1 & 16.2 & 14.7 & 7.4 & 7.7 \\
CVQA & 1 & 21.4 & 21.6 & 9.9 & 10.5 \\
ReasonVQA & 1 & 26.8 & 21.8 & 14.5 & 11.6 \\
MMStar & 1 & 18.8 & 18.8 & 7.2 & 8.1 \\
\bottomrule
\end{tabular}
}
\caption{Average number of VLM inference calls per question, computed as $K \times (\text{avg rounds} - 1)$ with $K=5$ (the final round is an answer commit with no VLM call); VLM Direct issues a single call per question. Reasoner: Qwen3.5-122B-A10B.}
\label{tab:vlm_calls_app}
\end{table}

\clearpage

\begin{table*}[!t]
\section{Full Results with Standard Deviations}
\label{sec:full-results}
\centering
\small
\setlength{\tabcolsep}{2pt}
\resizebox{\textwidth}{!}{

\begin{tabular}{l l
                cc cc cc
                cc cc cc}
\toprule
& &
\multicolumn{6}{c}{InternVL3.5-8B} &
\multicolumn{6}{c}{Qwen2.5-VL-7B} \\
\cmidrule(lr){3-8}\cmidrule(lr){9-14}
& &
\multicolumn{2}{c}{Perception} &
\multicolumn{2}{c}{Charts} &
\multicolumn{2}{c}{Knowledge} &
\multicolumn{2}{c}{Perception} &
\multicolumn{2}{c}{Charts} &
\multicolumn{2}{c}{Knowledge} \\
Reasoner & Strategy &
Acc $\uparrow$ & ECE $\downarrow$ &
Acc $\uparrow$ & ECE $\downarrow$ &
Acc $\uparrow$ & ECE $\downarrow$ &
Acc $\uparrow$ & ECE $\downarrow$ &
Acc $\uparrow$ & ECE $\downarrow$ &
Acc $\uparrow$ & ECE $\downarrow$ \\
\midrule
Human & - & 72.0 & - & 86.0 & - & 69.3 & - & 70.0 & - & 92.0 & - & 74.0 & -\\
\midrule
& BP & \textbf{49.0} & - & 32.0 & - & 60.7 & - & 42.0 & - & 29.0 & - & 61.3 & - \\
GPT-5 Mini & UoT & 46.0 & - & 36.0 & - & 50.0 & - & 39.0 & - & 30.0 & - & 58.0 & -\\
& ReAct & 36.0 & - & 46.0 & - & 56.7 & - & 42.0 & - & 35.0 & - & 63.0 & - \\
\midrule[0.2pt]
& BP & 48.6 {\scriptsize$\pm$1.52} & 0.27 & \textbf{59.8} {\scriptsize$\pm$8.76} & 0.14 & \textbf{61.0} {\scriptsize$\pm$1.80} & 0.12 & 50.8 {\scriptsize$\pm$3.03} & 0.22 & \textbf{60.6} {\scriptsize$\pm$6.15} & 0.18 & \textbf{67.7} {\scriptsize$\pm$2.07} & 0.09 \\
Qwen3.5-122B-A10B & UoT & 40.6 {\scriptsize$\pm$4.39} & 0.14 & 29.0 {\scriptsize$\pm$4.30} & 0.55 & 54.3 {\scriptsize$\pm$0.89} & 0.14 & 45.2 {\scriptsize$\pm$5.40} & 0.22 & 35.4 {\scriptsize$\pm$2.19} & 0.47 & 59.9 {\scriptsize$\pm$1.56} & 0.13\\
& ReAct & 46.0 {\scriptsize$\pm$3.67} & 0.32 & 46.2 {\scriptsize$\pm$8.41} & 0.31 & 60.7 {\scriptsize$\pm$2.79} & 0.14 & \textbf{53.6} {\scriptsize$\pm$6.19} & 0.20 & 45.0 {\scriptsize$\pm$4.18} & 0.39 & 65.9 {\scriptsize$\pm$2.06} & 0.12 \\
\midrule[0.2pt]
& BP & 43.0 {\scriptsize$\pm$6.67} & 0.29 & 36.2 {\scriptsize$\pm$4.66} & 0.49 & 49.3 {\scriptsize$\pm$1.86} & 0.18 & 49.6{\scriptsize$\pm$4.34} & 0.19 & 35.2 {\scriptsize$\pm$2.77} & 0.53 & 54.1 {\scriptsize$\pm$2.40} & 0.17 \\
GPT-oss-20B & UoT & 36.4 {\scriptsize$\pm$4.72} & 0.19 & 30.2 {\scriptsize$\pm$5.22} & 0.57 & 42.7 {\scriptsize$\pm$3.02} & 0.13 & 41.8 {\scriptsize$\pm$4.60} & 0.20 & 28.4 {\scriptsize$\pm$2.61} & 0.59 & 45.8 {\scriptsize$\pm$3.21} & 0.15 \\
& ReAct & 39.0 {\scriptsize$\pm$5.57} & 0.32 & 33.4 {\scriptsize$\pm$4.62} & 0.52 & 50.6 {\scriptsize$\pm$2.00} & 0.19 & 47.6 {\scriptsize$\pm$4.98} & 0.28 & 42.8 {\scriptsize$\pm$3.56} & 0.38 & 52.0 {\scriptsize$\pm$2.02} & 0.21 \\
\midrule[0.2pt]
& BP & 43.6 {\scriptsize$\pm$3.51} & 0.31 & 43.4 {\scriptsize$\pm$4.28} & 0.41 & 42.7 {\scriptsize$\pm$2.54} & 0.22 & 49.4 {\scriptsize$\pm$5.94} & 0.16 & 45.0 {\scriptsize$\pm$3.16} & 0.41 & 48.1 {\scriptsize$\pm$2.37} & 0.20 \\
Qwen2.5-72B & UoT & 39.2 {\scriptsize$\pm$2.95} & 0.27 & 28.4 {\scriptsize$\pm$5.55} & 0.52 & 38.9 {\scriptsize$\pm$3.28} & 0.23 & 41.8 {\scriptsize$\pm$5.31} & 0.29 & 31.2 {\scriptsize$\pm$2.86} & 0.51 & 45.0 {\scriptsize$\pm$1.87} & 0.18 \\
& ReAct & 44.0 {\scriptsize$\pm$4.18} & 0.33 & 38.4 {\scriptsize$\pm$6.69} & 0.45 & 46.7 {\scriptsize$\pm$2.03} & 0.21 & 46.6 {\scriptsize$\pm$2.61} & 0.34 & 47.0 {\scriptsize$\pm$6.44} & 0.32 & 48.9 {\scriptsize$\pm$2.05} & 0.21 \\
\midrule[0.2pt]
& BP & 39.8 {\scriptsize$\pm$5.07} & 0.23 & 38.2 {\scriptsize$\pm$4.97} & 0.34 & 43.2 {\scriptsize$\pm$2.18} & 0.16 & 43.8 {\scriptsize$\pm$5.36} & 0.14 & 40.8 {\scriptsize$\pm$3.03} & 0.30 & 45.0 {\scriptsize$\pm$1.94} & 0.20 \\
Qwen3-32B & UoT & 29.2 {\scriptsize$\pm$3.70} & 0.16 & 30.8 {\scriptsize$\pm$3.77} & 0.35 & 29.9 {\scriptsize$\pm$2.28} & 0.11 & 26.0 {\scriptsize$\pm$4.30} & 0.15 & 34.8 {\scriptsize$\pm$5.76} & 0.24 & 31.3 {\scriptsize$\pm$2.67} & 0.18 \\
& ReAct & 37.6 {\scriptsize$\pm$6.11} & 0.31 & 39.2 {\scriptsize$\pm$5.26} & 0.41 & 44.7 {\scriptsize$\pm$2.61} & 0.13 & 45.4 {\scriptsize$\pm$2.19} & 0.21 & 48.0 {\scriptsize$\pm$2.74} & 0.26 & 48.7 {\scriptsize$\pm$2.33} & 0.17 \\
\midrule[0.2pt]
& BP & 38.8 {\scriptsize$\pm$2.17} & 0.22 & 28.6 {\scriptsize$\pm$4.16} & 0.40 & 32.1 {\scriptsize$\pm$2.09} & 0.18 & 38.4 {\scriptsize$\pm$3.36} & 0.23 & 30.8 {\scriptsize$\pm$4.66} & 0.37 & 38.5 {\scriptsize$\pm$2.13} & 0.18 \\
Qwen3-8B & UoT & 27.0 {\scriptsize$\pm$2.55} & 0.20 & 19.6 {\scriptsize$\pm$3.44} & 0.42 & 24.3 {\scriptsize$\pm$3.26} & 0.15 & 27.4 {\scriptsize$\pm$3.71} & 0.23 & 20.6 {\scriptsize$\pm$2.70} & 0.45 & 26.3 {\scriptsize$\pm$1.68} & 0.18 \\
& ReAct & 38.2 {\scriptsize$\pm$1.30} & 0.31 & 25.8 {\scriptsize$\pm$3.63} & 0.60 & 36.5 {\scriptsize$\pm$1.54} & 0.22 & 46.0 {\scriptsize$\pm$3.54} & 0.19 & 29.4 {\scriptsize$\pm$2.07} & 0.57 & 39.5 {\scriptsize$\pm$2.19} & 0.24 \\
\midrule[0.2pt]
& BP & 38.0 {\scriptsize$\pm$4.00} & 0.32 & 34.8 {\scriptsize$\pm$6.46} & 0.25 & 32.9 {\scriptsize$\pm$3.08} & 0.28 & 42.8 {\scriptsize$\pm$3.70} & 0.19 & 29.2 {\scriptsize$\pm$3.19} & 0.28 & 35.7 {\scriptsize$\pm$2.44} & 0.21 \\
Qwen2.5-7B & UoT & 32.4 {\scriptsize$\pm$5.13} & 0.24 & 25.8 {\scriptsize$\pm$6.50} & 0.38 & 31.1 {\scriptsize$\pm$2.24} & 0.24 & 35.0 {\scriptsize$\pm$1.58} & 0.26 & 26.8 {\scriptsize$\pm$3.42} & 0.42 & 31.9 {\scriptsize$\pm$2.35} & 0.23 \\
& ReAct & 37.0 {\scriptsize$\pm$2.12} & 0.31 & 32.8 {\scriptsize$\pm$2.17} & 0.55 & 34.9 {\scriptsize$\pm$2.21} & 0.25 & 45.0 {\scriptsize$\pm$6.71} & 0.26 & 27.8 {\scriptsize$\pm$3.35} & 0.59 & 39.5 {\scriptsize$\pm$2.95} & 0.27 \\
\midrule
& BP & 33.8 {\scriptsize$\pm$7.16} & 0.28 & 42.2 {\scriptsize$\pm$5.72} & 0.27 & 32.7 {\scriptsize$\pm$3.16} & 0.20 & 36.8 {\scriptsize$\pm$2.28} & 0.29 & 40.8 {\scriptsize$\pm$3.83} & 0.25 & 37.3 {\scriptsize$\pm$2.18} & 0.16 \\
LLaMa3.1-8B & UoT & 32.2 {\scriptsize$\pm$3.63} & 0.25 & 32.6 {\scriptsize$\pm$5.41} & 0.47 & 32.9 {\scriptsize$\pm$2.40} & 0.16 & 34.8 {\scriptsize$\pm$1.92} & 0.20 & 32.4 {\scriptsize$\pm$4.34} & 0.45 & 30.1 {\scriptsize$\pm$3.73} & 0.23 \\
& ReAct & 38.0 {\scriptsize$\pm$2.92} & 0.28 & 51.4 {\scriptsize$\pm$4.62} & 0.17 & 34.3 {\scriptsize$\pm$2.92} & 0.18 & 46.4 {\scriptsize$\pm$2.41} & 0.18 & 58.2 {\scriptsize$\pm$1.79} & 0.16 & 36.3 {\scriptsize$\pm$2.25} & 0.22 \\
\midrule
VLM & - & 27.0 {\scriptsize$\pm$3.38} & 0.44 & 35.7 {\scriptsize$\pm$3.74} & 0.50 & 25.2 {\scriptsize$\pm$2.00} & 0.43 & 20.3 {\scriptsize$\pm$2.61} & 0.49 & 33.7 {\scriptsize$\pm$5.29} & 0.43 & 25.3 {\scriptsize$\pm$2.32} & 0.33 \\
\bottomrule
\end{tabular}
}
\caption{
Full results with standard deviations across trials. Accuracy and calibration on three task categories:
\textbf{Perception} (SeedBench), \textbf{Charts} (FigureQA),
and \textbf{Knowledge} (macro-average over MMStar, CVQA, and ReasonVQA).
}
\label{tab:category-results-full}
\end{table*}

\begin{table*}[!t]
\section{Knowledge Category Breakdown}
\label{sec: knowledge breakdown}
\centering
\small
\setlength{\tabcolsep}{2pt}       
\resizebox{\textwidth}{!}{
\begin{tabular}{l l
                cc cc cc
                cc cc cc}
\toprule
& &
\multicolumn{6}{c}{InternVL3.5-8B} &
\multicolumn{6}{c}{Qwen2.5-VL-7B} \\
\cmidrule(lr){3-8}\cmidrule(lr){9-14}
& &
\multicolumn{2}{c}{MMStar} &
\multicolumn{2}{c}{CVQA} &
\multicolumn{2}{c}{ReasonVQA} &
\multicolumn{2}{c}{MMStar} &
\multicolumn{2}{c}{CVQA} &
\multicolumn{2}{c}{ReasonVQA} \\
Reasoner & Strategy &
Acc $\uparrow$ & ECE $\downarrow$ &
Acc $\uparrow$ & ECE $\downarrow$ &
Acc $\uparrow$ & ECE $\downarrow$ &
Acc $\uparrow$ & ECE $\downarrow$ &
Acc $\uparrow$ & ECE $\downarrow$ &
Acc $\uparrow$ & ECE $\downarrow$ \\
\midrule
Human & - & 60.00 & - & 78.00 & - & 70.00 & - & 68.00 & - & 78.00 & - & 76.00 & -\\
\midrule
& BP & \textbf{52.00} & - & 61.00 & - & \textbf{69.00} & - & 51.00 & - & 53.00 & - & \textbf{80.00} & - \\
GPT-5 Mini & UoT & 39.00 & - & \textbf{63.00} & - & 48.00 & - & 43.00 & - & \textbf{59.00} & - & 72.00 & -\\
& ReAct & 49.00 & - & 56.00 & - & 65.00 & - & \textbf{53.00} & - & 58.00 & - & 78.00 & - \\
\midrule[0.2pt]
& BP & 50.6 {\scriptsize$\pm$2.70} & 0.20 & 61.2 {\scriptsize$\pm$2.86} & 0.09 & 71.2 {\scriptsize$\pm$1.92} & 0.05 & 61.0 {\scriptsize$\pm$3.54} & 0.13 & 61.0 {\scriptsize$\pm$4.64} & 0.10 & 81.2 {\scriptsize$\pm$3.27} & 0.05 \\
Qwen3.5-122B-A10B & UoT & 44.4 {\scriptsize$\pm$3.21} & 0.17 & 53.2 {\scriptsize$\pm$3.83} & 0.18 & 65.4 {\scriptsize$\pm$3.29} & 0.07 & 52.2 {\scriptsize$\pm$4.76} & 0.16 & 53.6 {\scriptsize$\pm$5.13} & 0.12 & 73.8 {\scriptsize$\pm$2.77} & 0.10 \\
& ReAct & 50.8 {\scriptsize$\pm$5.89} & 0.18 & 62.2 {\scriptsize$\pm$4.49} & 0.14 & 69.0 {\scriptsize$\pm$3.32} & 0.09 & 57.8 {\scriptsize$\pm$3.83} & 0.15 & 60.0 {\scriptsize$\pm$3.81} & 0.13 & 80.0 {\scriptsize$\pm$4.90} & 0.07 \\
\midrule[0.2pt]
GPT-oss-20B & BP & 47.60 {\scriptsize$\pm$4.28} & 0.18 & 40.00 {\scriptsize$\pm$2.55} & 0.26 & 60.20 {\scriptsize$\pm$2.49} & 0.09 & 51.40 {\scriptsize$\pm$3.36} & 0.17 & 40.80 {\scriptsize$\pm$4.32} & 0.25 & 70.00 {\scriptsize$\pm$4.69} & 0.09 \\
& UoT & 37.00 {\scriptsize$\pm$5.10} & 0.19 & 41.40 {\scriptsize$\pm$1.67} & 0.09 & 49.80 {\scriptsize$\pm$7.29} & 0.10 & 41.60 {\scriptsize$\pm$4.93} & 0.14 & 35.80 {\scriptsize$\pm$5.76} & 0.20 & 60.00 {\scriptsize$\pm$5.96} & 0.11 \\
& ReAct & 48.00 {\scriptsize$\pm$4.47} & 0.25 & 45.20 {\scriptsize$\pm$2.28} & 0.19 & 58.60 {\scriptsize$\pm$3.29} & 0.13 & 48.60 {\scriptsize$\pm$5.41} & 0.23 & 38.60 {\scriptsize$\pm$2.30} & 0.30 & 68.80 {\scriptsize$\pm$1.48} & 0.09 \\
\midrule[0.2pt]
Qwen2.5-72B & BP & 42.00 {\scriptsize$\pm$6.08} & 0.24 & 43.20 {\scriptsize$\pm$2.68} & 0.19 & 42.80 {\scriptsize$\pm$3.70} & 0.22 & 42.60 {\scriptsize$\pm$3.13} & 0.22 & 40.40 {\scriptsize$\pm$4.04} & 0.29 & 61.40 {\scriptsize$\pm$4.93} & 0.09 \\
& UoT & 37.60 {\scriptsize$\pm$6.35} & 0.20 & 39.00 {\scriptsize$\pm$5.10} & 0.28 & 40.20 {\scriptsize$\pm$5.54} & 0.21 & 41.60 {\scriptsize$\pm$2.30} & 0.21 & 41.00 {\scriptsize$\pm$3.32} & 0.22 & 52.40 {\scriptsize$\pm$3.91} & 0.11 \\
& ReAct & 45.40 {\scriptsize$\pm$2.51} & 0.25 & 48.80 {\scriptsize$\pm$3.27} & 0.18 & 46.00 {\scriptsize$\pm$4.47} & 0.22 & 45.00 {\scriptsize$\pm$3.32} & 0.24 & 37.60 {\scriptsize$\pm$2.07} & 0.31 & 64.20 {\scriptsize$\pm$4.76} & 0.09 \\
\midrule[0.2pt]
Qwen3-32B & BP & 41.40 {\scriptsize$\pm$4.16} & 0.19 & 41.60 {\scriptsize$\pm$3.91} & 0.17 & 46.60 {\scriptsize$\pm$3.21} & 0.13 & 42.00 {\scriptsize$\pm$2.55} & 0.19 & 34.80 {\scriptsize$\pm$2.59} & 0.24 & 58.20 {\scriptsize$\pm$4.55} & 0.16 \\
& UoT & 26.20 {\scriptsize$\pm$3.11} & 0.13 & 35.40 {\scriptsize$\pm$4.28} & 0.06 & 28.20 {\scriptsize$\pm$4.32} & 0.14 & 28.60 {\scriptsize$\pm$4.16} & 0.15 & 28.40 {\scriptsize$\pm$5.41} & 0.20 & 36.80 {\scriptsize$\pm$4.21} & 0.18 \\
& ReAct & 43.00 {\scriptsize$\pm$4.00} & 0.16 & 41.20 {\scriptsize$\pm$6.14} & 0.17 & 49.80 {\scriptsize$\pm$2.77} & 0.07 & 47.20 {\scriptsize$\pm$4.76} & 0.15 & 37.20 {\scriptsize$\pm$3.56} & 0.23 & 61.60 {\scriptsize$\pm$3.65} & 0.12 \\
\midrule[0.2pt]
Qwen3-8B & BP & 32.00 {\scriptsize$\pm$4.00} & 0.19 & 26.00 {\scriptsize$\pm$3.39} & 0.25 & 38.20 {\scriptsize$\pm$3.42} & 0.11 & 38.40 {\scriptsize$\pm$3.78} & 0.14 & 29.00 {\scriptsize$\pm$4.18} & 0.27 & 48.20 {\scriptsize$\pm$3.03} & 0.12 \\
& UoT & 23.00 {\scriptsize$\pm$6.67} & 0.19 & 25.20 {\scriptsize$\pm$5.40} & 0.11 & 24.80 {\scriptsize$\pm$4.66} & 0.16 & 27.60 {\scriptsize$\pm$1.95} & 0.12 & 22.80 {\scriptsize$\pm$3.63} & 0.21 & 28.60 {\scriptsize$\pm$2.88} & 0.22 \\
& ReAct & 40.00 {\scriptsize$\pm$3.46} & 0.20 & 29.40 {\scriptsize$\pm$1.52} & 0.31 & 40.20 {\scriptsize$\pm$2.68} & 0.16 & 36.80 {\scriptsize$\pm$2.17} & 0.32 & 33.60 {\scriptsize$\pm$4.93} & 0.27 & 48.20 {\scriptsize$\pm$3.77} & 0.12 \\
\midrule[0.2pt]
Qwen2.5-7B & BP & 32.80 {\scriptsize$\pm$6.53} & 0.33 & 29.20 {\scriptsize$\pm$6.14} & 0.32 & 36.80 {\scriptsize$\pm$2.28} & 0.20 & 32.80 {\scriptsize$\pm$4.09} & 0.21 & 30.60 {\scriptsize$\pm$4.72} & 0.32 & 43.80 {\scriptsize$\pm$3.83} & 0.09 \\
& UoT & 33.80 {\scriptsize$\pm$4.09} & 0.19 & 29.40 {\scriptsize$\pm$2.97} & 0.28 & 30.20 {\scriptsize$\pm$4.44} & 0.26 & 29.40 {\scriptsize$\pm$5.13} & 0.25 & 29.20 {\scriptsize$\pm$2.86} & 0.30 & 37.20 {\scriptsize$\pm$3.90} & 0.15 \\
& ReAct & 38.20 {\scriptsize$\pm$3.56} & 0.20 & 31.20 {\scriptsize$\pm$3.11} & 0.31 & 35.40 {\scriptsize$\pm$4.67} & 0.25 & 35.60 {\scriptsize$\pm$6.43} & 0.34 & 33.00 {\scriptsize$\pm$3.08} & 0.37 & 50.00 {\scriptsize$\pm$5.24} & 0.11 \\
\midrule[0.2pt]
& BP & 29.80 {\scriptsize$\pm$5.81} & 0.20 & 37.80 {\scriptsize$\pm$4.55} & 0.14 & 30.40 {\scriptsize$\pm$5.94} & 0.27 & 32.80 {\scriptsize$\pm$3.96} & 0.17 & 36.00 {\scriptsize$\pm$2.45} & 0.21 & 43.20 {\scriptsize$\pm$4.60} & 0.11 \\
LLaMa3.1-8B & UoT & 29.00 {\scriptsize$\pm$4.74} & 0.16 & 37.80 {\scriptsize$\pm$3.90} & 0.12 & 31.80 {\scriptsize$\pm$3.77} & 0.22 & 25.60 {\scriptsize$\pm$7.09} & 0.24 & 27.60 {\scriptsize$\pm$3.78} & 0.29 & 37.00 {\scriptsize$\pm$7.78} & 0.17 \\
& ReAct & 30.80 {\scriptsize$\pm$1.79} & 0.18 & 39.60 {\scriptsize$\pm$5.41} & 0.12 & 32.40 {\scriptsize$\pm$6.66} & 0.24 & 32.00 {\scriptsize$\pm$5.00} & 0.22 & 32.20 {\scriptsize$\pm$3.96} & 0.30 & 44.60 {\scriptsize$\pm$2.19} & 0.15 \\
\midrule[0.2pt]
VLM & - & 27.73 {\scriptsize$\pm$3.32} & 0.41 & 25.64 {\scriptsize$\pm$2.98} & 0.42 & 22.36 {\scriptsize$\pm$4.01} & 0.46 & 27.64 {\scriptsize$\pm$4.84} & 0.34 & 28.09 {\scriptsize$\pm$4.11} & 0.31 & 20.09 {\scriptsize$\pm$2.84} & 0.34 \\
\bottomrule
\end{tabular}
}
\caption{
Knowledge-based category breakdown showing accuracy and calibration on:
\textbf{MMStar}, \textbf{CVQA}, and \textbf{ReasonVQA}.
}
\label{tab:knowledge-breakdown}
\end{table*}

\clearpage

\section{Baseline Prompt}
\label{sec:baseline_prompt}
\label{sec: bp}
\begin{promptbox}{Belief-State Prompt Used to Evaluate BP method}

\paragraph{Belief Tracking}
\begin{itemize}
    \item Initialize: If 2 options $\to$ 50\% each; If 4 options $\to$ 25\% each
    \item Update beliefs after each question based on response patterns and the VLM's confidence
    \item Aim for $\ge$90\% confidence before finalizing your answer
    \item Only provide final answer when truly confident you've eliminated alternatives
    \item Responses with low confidence score will need to be further cross-validated because the VLM is providing conflicting information.
\end{itemize}

\paragraph{Output Format}
At every step, you must include the following and correct format:
\begin{itemize}
    \item \textbf{Thought:} Before every question or final answer, explicitly state your thought process by outputting `Thought: <complete description of your rationales>'.
    \item \textbf{Action:} Then output exactly one of:
    \begin{itemize}
        \item `My question is: <fully self-contained question>'
        \item `The answer is: (A)' or `(B)', `(C)', `(D)'
    \end{itemize}
\end{itemize}

\paragraph{Critical Format Rules}
\begin{itemize}
    \item Phrase ``My question is: '' must appear exactly when asking a question
    \item Phrase ``The answer is: '' must appear exactly when providing final answer
    \item Output only ONE thought \& ONE action per turn
    \item Each ``My question'' must include all necessary context (e.g., ``about the largest red shape'', ``regarding the texture of the object on the right'').
    \item You may ask multiple questions (up to the high interaction budget), so you should not rely too heavily on early questions.
\end{itemize}

\paragraph{Decision Points}
Ask another question when:
\begin{itemize}
    \item You're uncertain which option is correct
    \item You got conflicting answers that need clarification (The responses with low confidence score, such as 1/5 or 2/5)
    \item You want to verify your leading hypothesis
    \item You haven't yet tested all the key differences between options
\end{itemize}
Provide your final answer when:
\begin{itemize}
    \item One option is clearly supported by multiple reliable observations
    \item You've tested the main alternatives and ruled them out
    \item You are $\ge$90\% confident about the final option
    \item You've asked enough questions to feel justified in your conclusion
\end{itemize}

\paragraph{KEY REMINDERS}
\begin{itemize}
    \item Think about what you actually learned and what you're still uncertain about
    \item Ask follow-ups when you notice contradictions
    \item Make decisions when you feel reasonably certain
    \item Each question should be fully self-contained with all necessary context
\end{itemize}

\noindent\rule{\linewidth}{0.4pt}
Begin now with your first question. Think about what distinguishing features matter for this problem, and ask an exploratory question that will help you understand the scene.
\end{promptbox}
\label{fig:bp_prompt}

\clearpage

\label{sec: uot}
\begin{promptbox}{UoT: Prompt for Generating Candidate Questions In Visual Setting}
\small
You are helping a blind reasoner answer an MCQ about an image. You will generate \{n\_candidates\} diverse visual questions that could help distinguish between the MCQ options.

\textbf{MCQ Question:} \{mcq\_question\} \\
\textbf{Options:} \{options\} \\
\textbf{Round:} \{round\_num\}

\vspace{0.5em}
\textbf{Previous questions and answers:} \\
\{history\}

\vspace{0.5em}
Generate exactly \{n\_candidates\} diverse visual questions that could help distinguish between the MCQ options. Each question should:
\begin{enumerate}
    \item Be self-contained (include all necessary context)
    \item Target a specific visual feature that differs between options
    \item Help eliminate at least one MCQ option
\end{enumerate}

Output a JSON array of exactly \{n\_candidates\} question strings:
\begin{verbatim}
["Question 1 about specific visual feature", 
 "Question 2 about another feature", ...]
\end{verbatim}

Output ONLY the JSON array, no other text.
\end{promptbox}
\label{fig:uot_question_gen_prompt}

\begin{promptbox}{UoT: Prompt for Information Gain Estimation In Visual Setting}
\small
You are estimating how informative a visual question would be for solving an MCQ.

\textbf{MCQ Question:} \{mcq\_question\} \\
\textbf{Options:} \{options\} \\
\textbf{Current belief:} \{belief\_str\}

\vspace{0.5em}
\textbf{Candidate visual question:} ``\{candidate\_question\}''

\vspace{0.5em}
Imagine a vision model sees the image and answers this question.
Predict 2-3 most likely responses and how each would change your beliefs over the MCQ options.
Output ONLY the JSON object in the exact format below (no other text):

\begin{verbatim}
{
  "outcomes": [
    {
      "response": "likely answer 1", 
      "probability": 0.5, 
      "new_belief": {example_belief}
    },
    {
      "response": "likely answer 2", 
      "probability": 0.3, 
      "new_belief": {example_belief2}
    },
    {
      "response": "uncertain/unclear", 
      "probability": 0.2, 
      "new_belief": {current_belief}
    }
  ]
}
\end{verbatim}

\textbf{IMPORTANT:} 
\begin{itemize}
    \item Probabilities must sum to 1.0
    \item Each new\_belief must have exactly \{n\_options\} values summing to 1.0
    \item Output only valid JSON
\end{itemize}
\end{promptbox}
\label{fig:info_gain_prompt}

\clearpage

\label{sec: react}
\begin{promptbox}{ReAct: Vanilla Setup in Visual Setting}
\small

You are a text-only reasoner with no direct image access using the ReAct (Reasoning + Acting) framework to answer a multiple-choice question about an image. You interact with a Vision Language Model (VLM) that can see the image.

\paragraph{ReAct Framework\\}
ReAct combines explicit \textbf{reasoning traces} with \textbf{actions}. For each turn, you must:
\begin{enumerate}
    \item \textbf{Thought}: Explicitly reason about what you currently know, what information is missing, and what to do next
    \item \textbf{Action}: Either ask a visual question OR provide your final answer
\end{enumerate}

\paragraph{Output Format\\}
Every response MUST follow this exact structure:

\textbf{When you need visual information:}
\begin{verbatim}
Thought: <your reasoning about the current situation and what visual info would help> 
My question is: <your visual question - must be self-contained with all necessary context>
\end{verbatim}

\textbf{When you are ready to answer:}
\begin{verbatim}
Thought: <your reasoning summarizing what you learned and why you chose this answer> 
The answer is: (A/B/C/D/E)
\end{verbatim}

\paragraph{Critical Format Rules}
\begin{itemize}
    \item The phrase ``My question is: '' must appear exactly when asking a question
    \item The phrase ``The answer is: '' must appear exactly when providing final answer
    \item These phrases are how the pipeline recognizes your output—if they're missing or slightly different, the system will fail
    \item Output only ONE thought + ONE action per turn
    \item Each ``My question'' must include all necessary context (e.g., ``about the largest red shape,'' ``regarding the texture of the object on the right'').
\end{itemize}

\paragraph{VLM RESPONSES\\}
The VLM responses will include a confidence score (e.g., ``Confidence: 3/5'').
Use your Thought to evaluate responses critically and decide your next action.

\paragraph{EXAMPLE TRACE\\}
\textbf{Question:} What color is the car in the image?
(A) Red (B) Blue (C) Green (D) Yellow

\begin{verbatim}
Thought: I need to determine the color of the car to answer this question. I should 
directly ask about the car's color.
My question is: What is the color of the car visible in the image?
\end{verbatim}

\textbf{Response:} The car appears to be blue. \\
\textbf{Confidence:} 4/5

\begin{verbatim}
Thought: The VLM reports the car is blue with high confidence (4/5). This matches 
option (B). I have sufficient evidence.
The answer is: (B)
\end{verbatim}

\vspace{0.5em}
\noindent\rule{\linewidth}{0.4pt}
\vspace{0.5em}
Begin now. Think about what visual information would help distinguish between the answer options, then ask your first question.

\end{promptbox}
\label{fig:uot_system_prompt}

\begin{figure*}[t]

\section{Query Type Classification}
\label{sec:query_classify}

\noindent \textbf{Query Categories.} We classify each reasoner query into exactly one of five types:

\begin{itemize}
  \item \textbf{Descriptive} — Broad, open-ended request for a scene or image overview. The reasoner is building an initial mental model without targeting a specific hypothesis (e.g., \textit{``Describe the image''}, \textit{``What do you see?''}).

  \item \textbf{Targeted} — Narrow question about a specific visual attribute, object, or detail relevant to the MCQ options. The reasoner already has a direction and seeks focused evidence (e.g., \textit{``What color is the shirt?''}, \textit{``How many people are visible?''}).

  \item \textbf{Contrastive} — Rephrases or stress-tests a prior VLM answer to probe its stability. Often uses phrasing such as ``or'', ``actually'', ``are you sure'', or asks the same thing in a different way (e.g., \textit{``You said red---could it be orange?''}).

  \item \textbf{Confirmatory} — Directly asks the VLM to verify or agree with a specific claim or option. Unlike Contrastive, it seeks confirmation rather than challenge (e.g., \textit{``Can you confirm this is a department store?''}, \textit{``So the answer is (B), correct?''}).

  \item \textbf{Comparative} — Asks the VLM to rank or evaluate multiple MCQ options against the image simultaneously (e.g., \textit{``Between a mall and a department store, which fits better?''}).
\end{itemize}

\noindent\textbf{Pipeline.}
We classify 1{,}937 queries drawn from one trial each of human, Qwen3.5-122B-A10B (BP), and Qwen2.5-7B (BP) runs on SeedBench and CVQA (262, 225, 379, 469, 270, and 332 queries respectively). For each query, we provide the LLM judge with the MCQ question, the query itself, the round number, and the list of prior queries in that conversation. The judge outputs a one-sentence rationale followed by a structured JSON response with the assigned type and a confidence score.

\noindent\textbf{Validation.}
Following the same annotation proxy used for the semantic equivalence judge, three annotators independently labeled 200 stratified instances (100 per model: 40 per query type from Qwen3.5-122B-A10B and Qwen2.5-7B). Disagreements were resolved through discussion to form a consensus label, and an item is counted as correct if at least one annotator agrees with the judge. The judge achieves a union match rate of 82.5\% (165/200) and a Cohen's $\kappa$ of 0.781, confirming it as a reliable proxy for query-type annotation.

\end{figure*}

\begin{figure*}[t]

\section{Belief Probe Assessment}
\label{sec:belief_probe_prompt}

\noindent\textbf{Detailed Setup.}
For each stored BP conversation of length $T$ rounds, we replay the trajectory using an \emph{assessment prompt}, then ask the same LLM to output a probability distribution $b_t \in \Delta^{|\mathcal{O}|}$ over the MCQ options based only on the evidence available up to step $t$.

Concretely, let $h_1$ denote the MCQ question message, $h_{2t}$ the VLM response (with consistency score) at round $t$, and $h_{2t+1}$ the model's subsequent chain-of-thought. The context fed to the model for probe step $t$ ($1 \le t \le T-1$) is:
\begin{equation}
  [\,p_{\text{assess}},\; h_1,\; h_2, h_3,\;\dots,\; h_{2t}, h_{2t+1}\,],
\end{equation}
ending with the model's own chain-of-thought after processing the $t$-th VLM response. The initial belief $b_0$ is set to the uniform prior, which is hardcoded.

We ran this procedure over 2{,}000 trajectories across four settings (Qwen3.5-122B-A10B and Qwen2.5-7B on SeedBench and CVQA). The assessment prompt $p_{\text{assess}}$, shown below, replaces the original BP system prompt:

\begin{promptbox}{Assessment Prompt for Belief Probing}
\small
You are re-examining your own past reasoning on a Visual Question Answering (VQA) task. Imagine yourself as the reasoner at this exact point in the conversation---you have seen only the evidence gathered so far, and you have not yet seen what comes next. Based on the reasoning you expressed---your thoughts, interpretations, stated confidence or doubt---trace back what your implicit belief state was and express it as a probability distribution over the MCQ options.

Your response MUST end with a JSON object on its own line. Example: \texttt{\{``A'': 0.25, ``B'': 0.50, ``C'': 0.15, ``D'': 0.10\}}

Rules: Include ALL options; probabilities must sum to 1.0; be faithful to what the reasoning actually expressed---do not inject new analysis.
\end{promptbox}

\end{figure*}

\begin{figure*}[t]
\section{Accuracy and Consistency Correlation}
\label{sec: correlation_accuracy_consistency}
  \centering
  \includegraphics[width=0.95\textwidth]{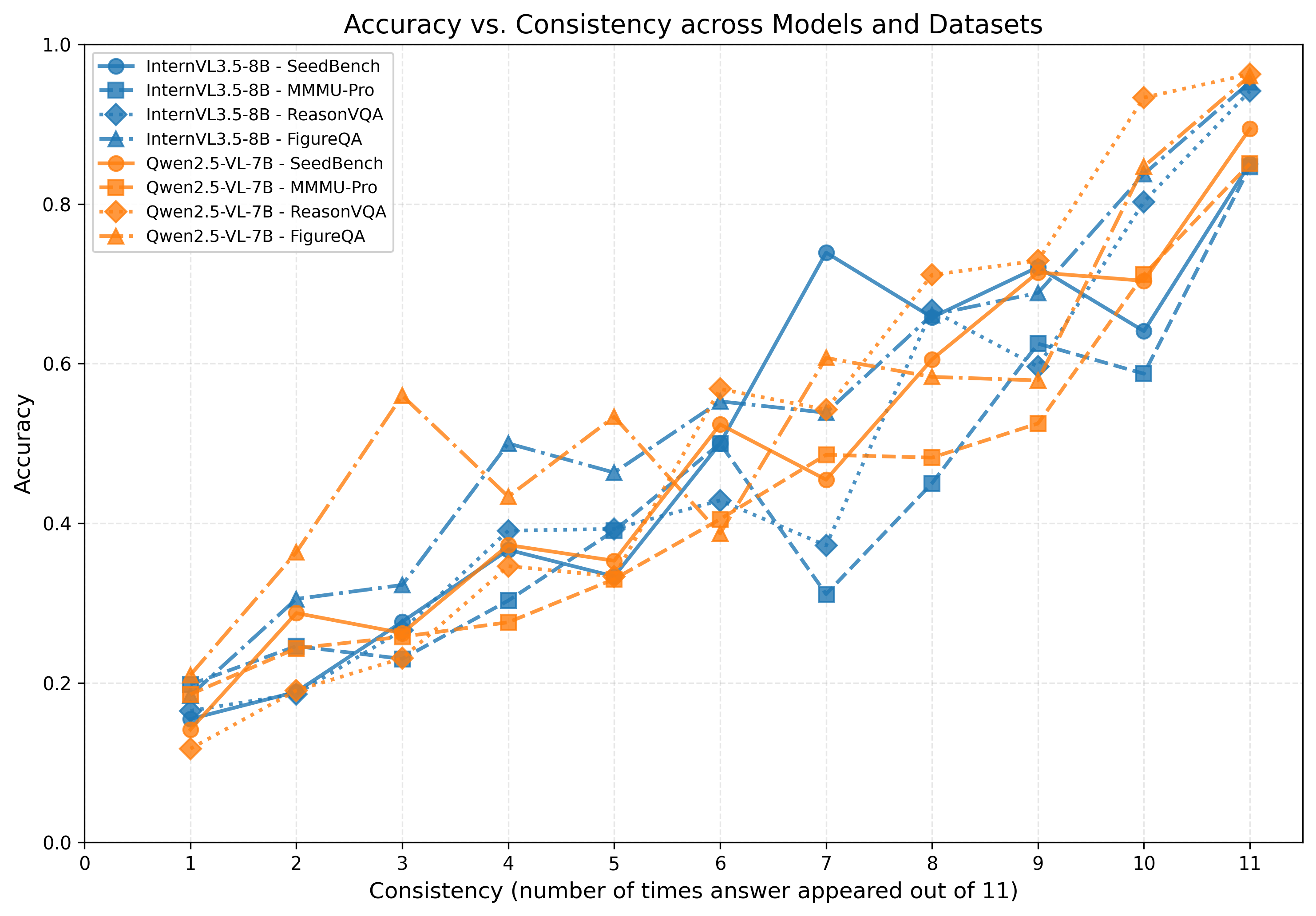}
  \caption{Accuracy vs. Consistency across different datasets for InternVL3.5-8B and Qwen2.5-VL-7B. Solid lines represent the base VLM performance, while dashed lines indicate reasoning performance. The x-axis represents the consistency level (1-11), showing that higher sensor consistency generally correlates with better downstream accuracy.}
  \label{fig:consistency}
\end{figure*}

\begin{figure*}[t]
\section{Expected Calibration Error (ECE)}
\label{sec: ece}
  \centering
  \includegraphics[width=0.95\textwidth]{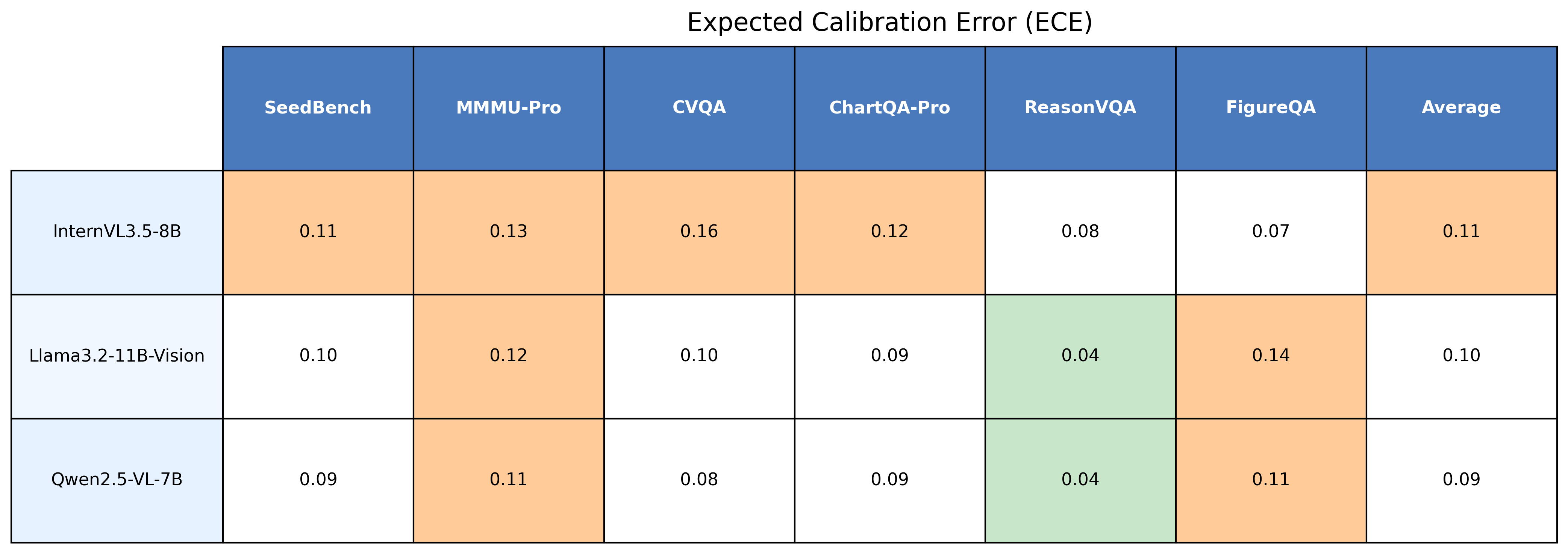}
  \caption{Preliminary effort to select the two most distinct VLMs (InternVL3.5-8B and Qwen2.5-VL-7B) based on the largest gap in Expected Calibration Error (ECE) across diverse datasets.}
  \label{fig:ece_preliminary}
\end{figure*}
\clearpage

    \section{Model Traces on Belief-State Prompting}
    \label{sec: traces}

\begin{figure}[h]
\centering
\includegraphics[width=0.38\columnwidth]{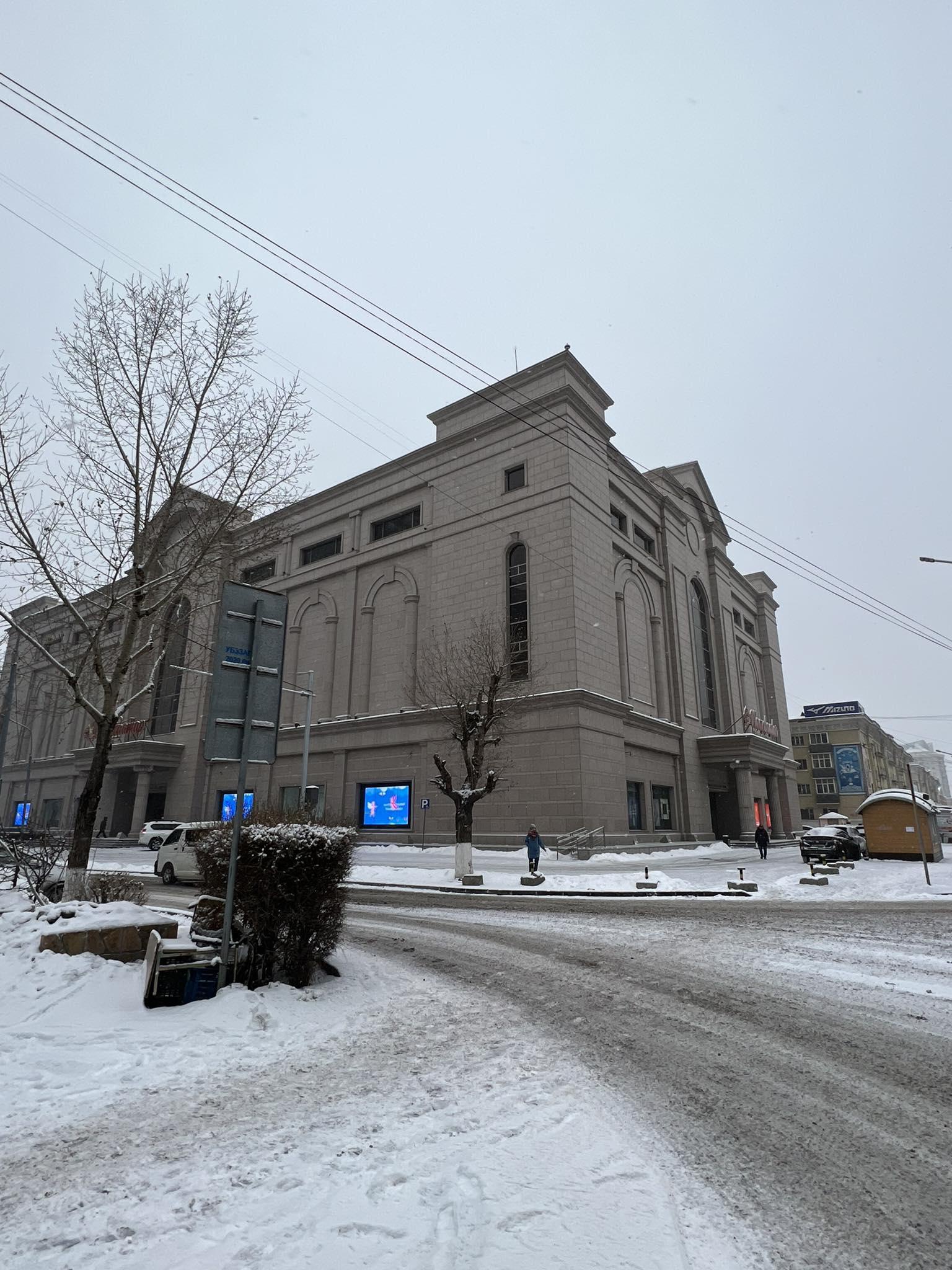}
\caption{CVQA instance: a large neoclassical building in Ulaanbaatar, Mongolia, photographed in winter. Ground truth: \textbf{(C) Ulaanbaatar Department Store}. (see App.~\ref{sec:error_analysis}).}
\label{fig:cvqa_ulaanbaatar}
\end{figure}

  \begingroup
  \footnotesize
  \textbf{Step 1}\\
  To determine the correct answer, I need to gather information about the building's characteristics, such as its name, location, and purpose. I'll start by asking about the building's name to see if the VLM can provide a direct answer.\\
  My question is: ``What is the name of the building in the image?''\\
  The building in the image appears to be a theater, characterized by its large entrance and classical architecture. Confidence: 3/5\\[3pt]

  \textbf{Step 2}\\
  The VLM provided a response that the building appears to be a theater, but this does not match any of the given options. The confidence level is moderate, so I need to ask more specific questions to clarify the building's identity. I will focus on the building's name and purpose to narrow down the options.\\
  My question is: ``Can you confirm the name of the building in the image, specifically if it is called Hunnu Mall, Department Store, Ulaanbaatar Department Store, or Sport Center?''\\
  The building in the image appears to be the Ulaanbaatar Department Store. Confidence: 4/5\\[3pt]

  \textbf{Step 3}\\
  The VLM has provided a more specific and relevant answer, indicating that the building is the Ulaanbaatar Department Store with a confidence level of 4/5. This is a strong indication, but I want to ensure there are no contradictions or ambiguities. I will ask a follow-up question to confirm the building's purpose to further validate this answer.\\
 My question is: ``Can you confirm if the building in the image is used as a department store?''\\
  The building appears to have an ornate facade with large arches, which is characteristic of historic architecture from certain European countries. It's less common for modern department stores to adopt such traditional architectural styles, suggesting it might serve another purpose or be a historical building, potentially not currently used as a department store. However, without more specific labels or signs visible on the building, it is difficult to confirm the exact function. Confidence: 3/5\\[3pt]

  \textbf{Step 4}\\
  The VLM's response introduces some uncertainty about the building's current use, despite the previous indication that it is the Ulaanbaatar Department Store. Given the moderate confidence level, I need to ask a more direct question to clarify the building's current function and name.\\
 My question is: ``Are there any signs, labels, or other indicators in the image that specifically identify the building as the Ulaanbaatar Department Store?''\\
  No, the image does not show any visible signage or labels that clearly identify the building as the Ulaanbaatar Department Store. Confidence: 4/5\\[3pt]

  \textbf{Step 5}\\
 The VLM's response indicates that there are no visible signs or labels identifying the building as the Ulaanbaatar Department Store, which introduces more uncertainty. However, the previous response with a 4/5 confidence level suggested it is the Ulaanbaatar Department Store. I will ask a final clarifying question to see if there are any other distinguishing features or details that can help confirm the building's identity.\\ 
  My question is: ``Can you describe any distinctive features or details of the building that might help identify it as one of the options: Hunnu Mall, Department Store, Ulaanbaatar Department Store, or Sport Center?''\\
  The building has a modern design with large windows, a flat roof, and decorative arches above the entryway. These features align more closely with those typically found in a shopping center like a mall or department store rather than a traditional sport center or single-use building. The presence of what appears to be advertisement screens on its facade further supports this conclusion. However, without more specific signage or interior clues, it's challenging to definitively classify the building as one of the given options. Confidence: 4/5\\[3pt]

  \textbf{Step 6}\\
  The VLM's response provides more details about the building's modern design and features, which align more closely with a shopping center like a mall or department store. The presence of advertisement screens further supports this. Given the consistent indication that the building is likely a department store and the lack of evidence pointing to the other options, I am now reasonably confident that the building is the Ulaanbaatar Department Store. The correct answer is (C) Ulaanbaatar Department Store\\
  The answer is: (C)

\endgroup

\section{Use of AI Tools}
\label{sec: ai statement}
We used AI-assisted tools (e.g., large language models) for grammatical and stylistic editing, and to help search for and organize relevant background information and literature. All technical content, experimental designs, analyses, and conclusions were created and verified by the authors, who remain responsible for any remaining errors.

\section{Annotation Instruction}
We recruited five expert annotators with prior experience in vision-and-language tasks. Annotators were instructed to carefully read each \method{} question and its answer options, and to use common sense and domain knowledge to select the single best answer. We first familiarized annotators with the task via a small pilot set of examples, discussed ambiguous cases, and provided feedback and clarifications on the guidelines before they proceeded to the final annotations. In knowledge-heavy domains, annotators were able to utilize external search engines and AI tools. In contrast, they were not allowed to access these tools in Perception and Chart domains.

\noindent \textbf{Annotation protocol and inter-annotator reliability.} To validate annotation quality before the full annotation effort, two annotators independently completed the first 50 questions of this benchmark spanning all five datasets. Any disagreements were discussed to clarify the guidelines, and each instance was successfully done by at least one of the two annotators. Following this validation, the remaining questions were distributed so that each question was handled by a single annotator, a design chosen deliberately for efficiency given the consistent pilot behavior. A question is retained if the annotator answered it correctly under the blind VLM interaction protocol.

\begin{figure*}[h!]
\section{Annotation Interface}
\label{sec:interface}
  \centering
  \includegraphics[width=\textwidth]{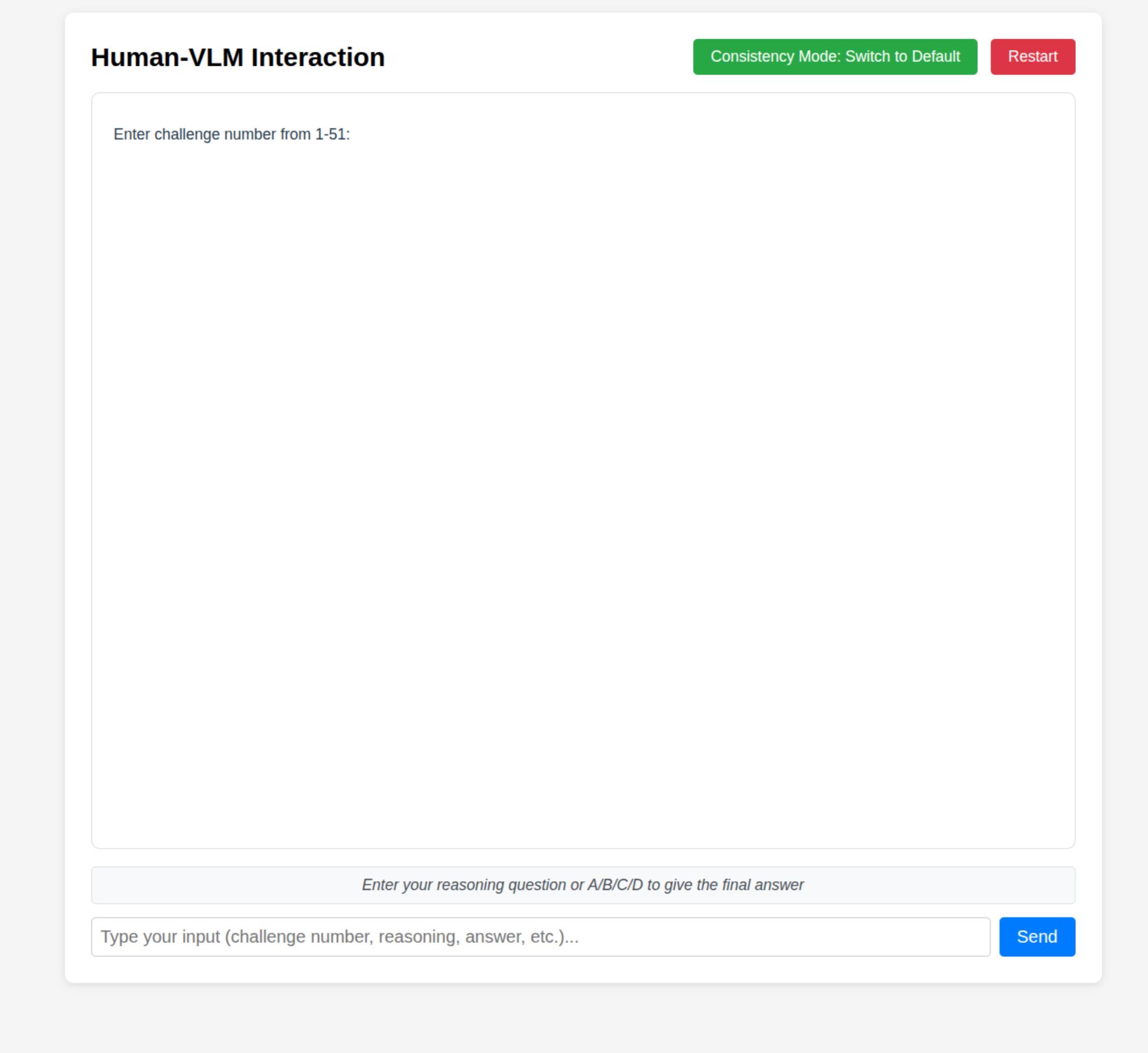}
  \caption{Annotation Interface for Human Annotators}
  \label{fig:annotation_interface}
\end{figure*}
\end{document}